\documentclass[preprint]{elsarticle}

\journal{Journal of Artificial Intelligence}

\AtBeginDocument{\providecommand\BibTeX{{\normalfont B\kern-0.5em{\scshape i\kern-0.25em b}\kern-0.8em\TeX}}}

\usepackage{amsmath}
\usepackage{mathtools}
\usepackage{microtype}        
\usepackage{ccicons}          
\usepackage{amsfonts}
\usepackage[official]{eurosym}
\usepackage{algorithm}
\usepackage{algorithmicx}
\usepackage[noend]{algpseudocode}
\algnewcommand{\LineComment}[1]{\Statex \(\triangleright\) #1}
\algnewcommand\algorithmicinput{\textbf{Input:}}
\algnewcommand\algorithmicoutput{\textbf{Output:}}
\algnewcommand\Input{\item[\algorithmicinput]}
\algnewcommand\Output{\item[\algorithmicoutput]}
\newcommand{\Break}{\State \textbf{break} }
\def\algbackskip{\hskip-\ALG@thistlm}
\usepackage{booktabs}
\usepackage{caption}
\usepackage{multirow}
\usepackage{subcaption}
\usepackage{array}
\usepackage[export]{adjustbox}
\usepackage{menukeys}
\usepackage[pdflang={en-US},pdftex]{hyperref}
\usepackage[section]{placeins}

\newtheorem{definition}{Definition}

\newcommand{\True}{{\texttt{True}}}
\newcommand{\False}{{\texttt{False}}}
\newcommand{\ltg}{{\textit{Local to Global}}}
\newcommand{\ethica}{{\scshape GLocalX}}
\newcommand{\corels}{{\scshape corels}}
\newcommand{\decisionsets}{{\emph{Decision Sets}}}
\newcommand{\lime}{\textsc{lime}}
\newcommand{\lore}{\textsc{lore}}
\newcommand{\shap}{\textsc{shap}}
\newcommand{\anchors}{\textsc{anchors}}
\newcommand{\cpar}{\textsc{cpar}}

\newcommand{\intrees}{\textsc{InTrees}}
\newcommand{\trepan}{\textsc{Trepan}}
\newcommand{\quinlan}{\textsc{C4.5}}
\newcommand{\xte}{$X_{le}$}
\newcommand{\qp}{quasi-polyhedron}
\newcommand{\join}{{\textsc{join}}}
\newcommand{\cut}{{\textsc{cut}}}
\newcommand{\code}[1]{\texttt{#1}}
\newcommand{\adult}{{\texttt{adult}}}
\newcommand{\compas}{{\texttt{compas}}}
\newcommand{\german}{{\texttt{german}}}
\newcommand{\diva}{{\texttt{diva}}}

\begin{document}

\begin{frontmatter}

\title{GLocalX - From Local to Global Explanations\\ of Black Box AI Models}

\author[unipi]{Mattia Setzu\corref{cor1}}
\ead{mattia.setzu@phd.unipi.it}
\cortext[cor1]{Corresponding author.}

\author[unipi]{Riccardo Guidotti}
\ead{riccardo.guidotti@unipi.it}

\author[unipi]{Anna Monreale}
\ead{anna.monreale@unipi.it}

\author[unipi]{Franco Turini}
\ead{turini@di.unipi.it}

\author[unipi]{\\Dino Pedreschi}
\ead{dino.pedreschi@unipi.it}

\author[isti]{Fosca Giannotti}
\ead{fosca.giannotti@isti.cnr.it}

\address[unipi]{University of Pisa, Largo B. Pontecorvo, Pisa, Italy}
\address[isti]{ISTI-CNR, Via G. Moruzzi, Pisa, Italy}

\begin{abstract}
Artificial Intelligence (AI) has come to prominence as one of the major components of our society, with applications in most aspects of our lives.
In this field, complex and highly nonlinear machine learning  models such as ensemble models, deep neural networks, and Support Vector Machines have consistently shown remarkable accuracy in solving complex tasks.
Although accurate, AI models often are ``black boxes'' which we are not able to understand.
Relying on these models has a multifaceted impact and raises significant concerns about their transparency.
Applications in sensitive and critical domains are a strong motivational factor in trying to understand the behavior of black boxes.
We propose to address this issue by providing an interpretable layer on top of black box models by aggregating \textit{``local''} explanations.
We present \ethica, a \textit{``local-first''} model agnostic explanation method.
Starting from local explanations expressed in form of local decision rules, \ethica\ iteratively generalizes them into global explanations by hierarchically aggregating them.
Our goal is to learn accurate yet simple interpretable models to emulate the given black box, and, if possible, replace it entirely.
We validate \ethica\ in a set of experiments in standard and constrained settings with limited or no access to either data or local explanations.
Experiments show that \ethica\ is able to accurately emulate several models with simple and small models, reaching state-of-the-art performance against natively global solutions. Our findings show how it is often possible to achieve a high level of both accuracy and comprehensibility of classification models, even in complex domains with high-dimensional data, without necessarily trading one property for the other. This is a key requirement for a trustworthy AI, necessary for adoption in high-stakes decision making applications.
\end{abstract}


\end{frontmatter}

\section{Introduction}
In the last decade, Artificial Intelligence (AI) decision systems have been widely used in a plethora of applications such as credit score, insurance risk, and health monitoring, in which accuracy is of the utmost importance~\cite{pasquale2015black}.
Complex nonlinear machine learning models such as ensemble models, deep neural networks (DNN) and Support Vector Machines (SVM) have shown remarkable performance in these tasks, and have made their way into a large number of systems~\cite{miller2019explanation}.
Unfortunately, their state-of-the-art performance comes at the cost of a clear interpretation of their inner workings~\cite{freitas2014comprehensible}.
These ``black box'' models have an opaque, hidden internal structure that humans do not understand~\cite{guidotti2018survey}.
Relying on black box systems is becoming increasingly risky both for their lack of transparency and the systematic bias they have shown in real-world scenarios~\cite{propublica2013analysis}.
The lack of proper explanations also has ethical implications, legally reported in the \textit{General Data Protection Regulation} (GDPR), approved by the European Parliament in May 2018.
The GDPR provides restrictions and guidelines for automated black box decision-making processes which, for the first time, introduces a ``right to explanation'' on the decisions of the system.
More specifically, the GDPR introduces a right to meaningful explanations when one is subject to automated AI systems~\cite{wachter2017right,comande2017right,guidotti2018helping}.

Given the great interest on the topic~\cite{goebel2018explainable,miller2019explanation,guidotti2018survey,adadi2018peeking}, several works in the literature try to explain opaque models with one of two goals: either providing instance explanations for a given decision by using a \textit{local} approach~\cite{ribeiro2016should,ribeiro2018anchors,guidotti2018local,panigutti2019marlena}, or providing \textit{global} explanations able to describe the overall logic of the black box~\cite{craven1996extracting,deng2014interpreting}.
These approaches differ both on their task and their use case, with local approaches having the upper hand in several scenarios.
As stated, local approaches~\cite{ribeiro2016should,lundberg2017unified} provide explanations of single instances and are beneficial to a plethora of users with different needs and resources at their disposal.
Model developers, who have access both to data and black box, can directly inspect the model; analysts can audit the model on a small sample of instances; users can retrieve an explanation on a decision that involves them directly, gain trust in the model and, possibly, an actionable recourse to follow up the decision.
On the other hand, global approaches have stricter requirements.
Users may require access both to the model and the data used to train and validate the model~\cite{yin2003cpar,craven1996extracting,lakkaraju2016interpretable}.
On the other hand, most local approaches are \textit{model agnostic}~\cite{ribeiro2016should,lundberg2017unified}, i.e., they do not assume knowledge of the black box or its internals, while global approaches may depend on the black box model~\cite{guidotti2018survey}.
In spite of this additional layer of abstraction, local approaches have repeatedly shown competitive performances similar to global approaches, mainly due to the smaller set of decisions to explain.
Finally, local approaches inherently enjoy a high degree of explanation independence, as they provide different views on data and are able to grasp qualitatively different patterns, providing different explanations for similar instances when confounding factors are at play~\cite{ross2018learning,ross2019ensembles}.
This behavior is reminiscent of \emph{bagging} estimators, as local approaches are a peculiar bagging in which each model is fit on one sample.
Feature transformations are also typical of local approaches~\cite{ribeiro2016should,guidotti2018local} as they are for bagging approaches.

Following these premises, we posit that leveraging \emph{local} approaches for tackling \emph{global} tasks can yield the benefits of the former and overcome the constraints of the latter.
In this paper, we decline the local-global dichotomy in favor of a \ltg\ interpretation and propose a formal definition of this problem and an algorithm to solve it.
Three assumptions~\cite{pedreschi2019meaningful} underpin our 
approach:
\begin{enumerate}
\item \textbf{Logical Explanability.} We believe that the cognitive vehicle for offering explanations should be close to the language of reasoning, that is \emph{logic}.
For this reason we adopt rule-based local explanations~\cite{ribeiro2018anchors,guidotti2019factual}.

\item \textbf{Local Explainability.} While a black box can be arbitrarily complex, we assume that in the neighborhood of each specific instance there is a high chance that the decision boundary of the black box is simple enough to be accurately approximated by an explanation~\cite{ribeiro2016should}.

\item \textbf{Explanation Composition.} We assume that similar instances admit similar explanations~\cite{alvarez2018robustness,guidotti2019stability} and that similar explanations are likely to be composed together into slightly more general ones.
\end{enumerate}

We support these hypotheses with \ethica\ (\textsc{GLO}bal to lo\textsc{CAL} e\textsc{X}plainer), a model-agnostic \textit{``local-first''} explanation algorithm. 
\ethica\ is based on the idea of deriving global explanations by inference on a set of logical rules representing local explanations. 
\ethica\ aggregates local explanations expressed in form of logical rules into a global explanation by iteratively ``merging'' the rules in a growing hierarchical structure while accounting for both fidelity, i.e., accuracy in emulating black box predictions, and complexity of the rules.
The merge procedure estimates a distance between explanations and yields a set of sorted candidate pairs to merge.
Then, the pair with minimum distance satisfying constraints on both fidelity and complexity is processed to guarantee generalization and the updated explanation replaces the selected pairs.
Constraint satisfaction is ensured on each merge.
This guarantees high fidelity and low complexity on the final explanation yielded by \ethica.

We showcase the \ltg\ formulation in two constrained scenarios typical of real-world use cases: in the former, we consider as available input a restricted number of local explanation rules; while in the latter, we consider no data available for the global explanation construction.
These settings can occur when the model is proprietary or data is inaccessible due to privacy concerns.
Empirical results over different black box models and datasets indicate that \ethica\ achieves both a high \textit{fidelity} and a low \textit{complexity} of the rule set representing the model explanation.
Compared to transparent models that either optimize model complexity or fidelity, but not both, \ethica\ reaches simultaneously high fidelity and low complexity.
The high accuracy in prediction tasks also suggests that \ethica\ might be used directly as a transparent model to replace global classifiers adopted in AI systems.

\smallskip
The rest of the paper is organized as follows.
Section ~\ref{sec:related_work} discusses related work.
Section~\ref{sec:problem} presents the local to global problem formulation and the idea adopted for solving it. 
Section~\ref{sec:ethica} describes the proposed hierarchical approach for the global explanation.
Experimental results are presented in  Section~\ref{sec:experiments}.
Finally, Section~\ref{sec:conclusions} concludes the paper and discusses new research directions.

\section{Related Work}
\label{sec:related_work}
To the best of our knowledge, no global frameworks merging local explanations are present in the literature.
However, related work comprises of a set of both \textit{global} and \textit{local} explanation methods~\cite{guidotti2018survey}.
In addition, we can distinguish between methods that explain the black box \emph{post-hoc}, and methods that are, on the other hand, \textit{explainable by design}~\cite{pedreschi2019meaningful}.

Explainable by design algorithms directly solve the classification problem~\cite{guidotti2018survey} and yield an interpretable and global model.
The most well known interpretable by design families of models comprise of the \textit{Decision Tree}~\cite{quinlan1993c4} (\textsc{dt}) and the rule-based classifiers (\textsc{rbc}), such as \cpar~\cite{yoon2012classification}, \decisionsets~\cite{schmidhuber2015deep,lakkaraju2017interpretable} and \corels~\cite{angelino2017learning}.
\cpar~\cite{yoon2012classification} combines the positive aspects of both associative classification and traditional rule-based classification, greedily generating a small set of rules directly from training data.
\decisionsets~\cite{lakkaraju2017interpretable} are ``collections of if-then rules that can be considered in any order''.
Given a set of rules, a set of functions is used to identify a subset which enjoys low internal conflict and several other desirable properties such as coverage, shortness, and accuracy.
\corels~\cite{angelino2017learning} also generates compact explanations with a optimality guarantees by using a discrete optimization technique for building sorted \textit{rule lists} over categorical feature spaces.
We observe different approaches to tackling the complexity of the interpretable global model, with some families ignoring it~\cite{yoon2012classification}, partially acknowledging it~\cite{ruggieri2004yadt} or fully including it~\cite{schmidhuber2015deep} in the construction phase.

Concerning post-hoc black box explanation, several foundational works rely on training explainable by design models leveraging queries to black box models.
An example of this family is the \emph{Conj Rules} approach~\cite{craven1994using}, which interprets a neural network by framing rule extraction as a learning problem.
Along the same lines is \trepan~\cite{craven1996extracting}, a refinement of the previous approach based on a decision tree specifically built for explaining the behavior of a black box.
The original training data and a randomized extension of it are labeled by the black box and used as training data for an interpretable model.
The model learns a Decision Tree that maximizes both the gain ratio~\cite{tan2007introduction} and the current model fidelity to the black box.
An advantage of \trepan\ with respect to common tree classifiers~\cite{quinlan1993c4} is that by enriching the dataset all splits are performed on a considerable amount of data.
Another approach using a single tree approximation as an interpretable explanation of the global model is \intrees~\cite{deng2014interpreting}.
It extracts, measures, prunes and selects the final explanation rules from tree ensembles, and calculates frequent variable interactions. 
We underline that the \intrees\ approach is model-dependent and cannot be generalized for explaining every black box, while in \trepan\ we can plug in any query-able black box, making \trepan\ a more powerful and flexible algorithm.

More recently, we are observing a shift of paradigm towards \textit{local} approaches for explaining the decision of black box classifiers for a single instance.
These approaches assume that complex black box models such as deep neural networks implement an overall global logic too complex to be explained and/or understood.
Conversely, they assume that local predictions of a single instance can be explained.
We call this assumption of explainability in a local neighborhood the \emph{local explainability} assumption.
\lime~\cite{ribeiro2016should} is the first attempt to derive a local explanation, as it relies on instances randomly generated in the neighborhood of the instance to be explained.
The authors propose a feature importance framework in which instances are mapped to a simpler interpretable space on which a \textit{linear model} is used to compute an importance score for each interpretable feature.
Interpretable features are then mapped back to the original feature space to provide an explanation.
\lime\ also provides a global feature importance tool to assess the feature relevance to discriminate in classification tasks, \textsc{LIME-SP}. 
Feature importance is the target of another local explainability model, \shap~\cite{DBLP:conf/nips/LundbergL17}, which frames feature importance as a collaborative game in which features are rewarded according to their contribution to the black box prediction.
The game is framed in a formal game theory setting in which features approximate the provably unique Shapley values~\cite{shapley1953value}.
Based on the neighborhood generation premise is also \lore~\cite{guidotti2018local}, which populates neighborhoods via genetic programming, optimizing both for the neighborhood distance and its label distribution.
As in \lime, the neighborhood is then used to train an explainable model from which an explanation is extracted.
\lore\ employs Decision Trees as explainable models, hence, other than returning a rule as an explanation, it is able to generate a set of counterfactual explanations, i.e., a set of rules similar to the one returned, but with a different outcome.
This feature is particularly valuable in actionable settings, in which the model user may understand what changes to apply to data to comply with the black box predictions.
The authors of \lime\ also propose \anchors~\cite{ribeiro2018anchors}, an algorithm that generates explanations in the form of decision rules by iteratively guessing premises and optimizing their precision.
An anchor explanation consists of a minimal set of premises that guarantees a baseline of accuracy even when new premises are added.
To quote the authors, an anchor ``is a rule [..] such that changes to the rest of the feature values of the instance do not matter.''
Local explanation methods have repeatedly shown high accuracy, seldom outperforming global models.
This, jointly with the \emph{local explainability} assumption, prompts us to ask whether we can leverage \textit{local} models to learn \textit{global} ones while preserving their high fidelity.

\section{Local to Global Explanation Problem}
\label{sec:problem}
In the following we introduce basic notations of classification on tabular data and we define the notion of \textit{explanation} and the \textit{local to global explanation problem} for which we propose a solution. 

A classifier is a function $f:\mathcal{X}^{(m)} \rightarrow \mathcal{Y}$ which maps data instances (tuples) $x$ from a feature space  $\mathcal{X}^{(m)}$ with $m$ input features to a decision $y$ in a label space $\mathcal{Y}$. 
We write $f(x) = y$ to denote the decision $y$ given by $f$, and $f(X) = Y$ as a shorthand for $\langle f(x) \ |\ x \in X \rangle$. 
We assume that any classifier can be queried at will.
Here we restrict to binary classification but the formulation and the solution can be easily extended to multi-class and multi-label problems. 
An instance $x$ consists of a set of $m$ attribute-value pairs $(a_i, v_i)$, where $a_i$ is a categorical or continuous feature and $v_i$ is a value from the domain of $a_i$. 
We denote with $b$ a \emph{black box} classifier whose internals are either unknown to the observer or known but uninterpretable by humans.
Examples include deep neural networks, SVMs, and ensemble classifiers like Random Forest and AdaBoost classifiers~\cite{tan2007introduction,miller2019explanation,guidotti2018survey}.

As \textit{explanation} $e$ we consider a decision rule $r$, i.e., $e = \langle r = P \rightarrow y \rangle$.
The decision rule $r$ describes the reason for the decision $y = b(x)$, $P = \{p_1, \dots, p_s\}$ is a set of \textit{premises} in conjunctive form, and $y$ is the rule \textit{outcome}.
As an example, let us consider the following explanation for a loan request for a user $x = \{ (\text{age}{=}22), (\text{job} = unemployed), (\text{amount}{ = }\text{ 10k}), (\text{car }{ = } \text{ no})\}$:
\begin{align*}
e = \langle & r = \{\text{age}{\geq}25, \text{ job }{ = }\text{ unemployed}, \text{ amount }{\leq}\text{ 10k}\} {\rightarrow} \text{deny} \rangle
\end{align*}
We name \textit{explanation theory} $E = \{ e_1, \dots, e_n\}$ a set of explanations, i.e., a set of decision rules.
We indicate with $\mathbb{E} = \{ E_1, \dots, E_N\}$ a set of explanation theories.
We account for logical explanation theories as the explanations are expressed in the form of logical rules.

According to~\cite{guidotti2018survey}, the \textit{local explanation problem} consists in retrieving an explanation which describes the reason behind a decision taken by a black box model $b$ for a single instance $x$ (local).
On the other hand, the \emph{global explanation problem} consists in finding the reasons for the classification for any instance in $X$ taken by a black box model $b$.
In this setting, the \textit{local to global problem} consists in exploiting a set of local explanations, describing the reasons behind single (local) decisions, to understand the overall  (global) logic of an opaque classifier used in an AI system.
Formally, \mbox{we define the problem as follows:}
%
%
\begin{definition}[Local to Global Explanation Problem]
    Let $e_1, \dots, e_n \in \mathcal{E}$ be a set of local explanations for a black box classifier $b$ defined in a human-interpretable domain $\mathcal{E}$. The \emph{local to global explanation problem} consists in finding a function $g$ yielding an \textit{explanation theory} $E = g(e_1, \dots, e_n) \subset \mathcal{E}$, such that $E$ describes the overall logic according to which $b$ makes decisions.
\end{definition}

In order to solve the local to global explanation  problem we need to generalize the local explanations, as they are accurate and faithful \emph{locally} but not \emph{globally}.
Given a black box $b$ adopted in an AI system, and a set of instances $X_{le} = \{x_1, \dots, x_n\}$ explained locally, and their local explanations $\{e_1, \dots, e_n\}$, we aim to solve the problem by deriving an explanation theory $E = \{ e'_1, \dots, e'_k\}$ by refining with an aggregation function $g$ the local explanations into an explanation theory emulating the global decision logic of the black box $b$.
Thus, the human-interpretable domain $\mathcal{E}$ consists in a set of logical decision rules.

\section{Local to Global Hierarchy of Explanation Theories}
\label{sec:ethica}
\ethica\ (\textsc{GLO}bal to lo\textsc{CAL} e\textsc{X}plainer) is an explanation method that hierarchically merges local explanations into a global explanation theory. 
In particular, \ethica\ takes as input a set of local explanations in form of explanation theories $\mathbb{E} = \{ E_1, \dots, E_n\}$ where each theory $E_i = \{e_i\}$ is formed by a single explanation, i.e., $|E_i| = 1 \; \forall E_i \in \mathbb{E}$.
\ethica\ iteratively \textit{merges} the explanation theories and finally returns an explanation theory $E = \{ e'_1, \dots, e'_k\}$ which emulates the global behavior of the black box $b$ simultaneously maintaining the overall model simple and interpretable.

At each iteration, \ethica\ merges the \textit{closest} pair of explanation theories $E_i, E_j$ by using a notion of \textit{similarity} between logical theories.
The pairs are filtered out according to merge quality criterion: if no pair satisfying the criterion is found, \ethica\ halts prematurely without building the full hierarchy.
The resulting hierarchy of explanation theories can be represented by using a tree-like diagram called \textit{dendrogram}~\cite{rokach2005clustering}.
There are two key elements in the \ethica\ approach: \emph{(i)} \emph{similarity search}, which allows to select which theories to merge and refine, and \emph{(ii)} a \emph{merge function}, which allows to refine the explanations.
We explore our choices in Section~\ref{sec:ethica_similarity} and Section~\ref{sec:ethica_merge}, respectively.
Finally Section~\ref{sec:classification} describes how to use for a classification task an explanation theory capturing the global behavior of the black box model. 

\begin{algorithm}[t]
\begin{algorithmic}[1]
  \Input $\mathbb{E}$ explanation theories, $\alpha$ filter threshold
  \Output $E$ explanation theory 
  
  \State E $\gets \emptyset$ 
  \Repeat
    \State $\mathbb{Q}$ $\gets$ \Call{sort}{$\mathbb{E}$} \Comment{\texttt{\scriptsize sort pairs of theories by similarity}}
    \State \textit{merged} $\gets$ \False
    \State $X' \gets$ batch(X)

    \While{$\neg$ \textit{merged} $\wedge$ $\mathbb{Q} \neq \emptyset$}
      \State $E_i, E_j \gets$ \Call{pop}{$\mathbb{Q}$}  \Comment{\texttt{\scriptsize select most similar theories}}
      \State $E_{i + j} \gets$ \Call{merge}{$E_i$, $E_j$, $X'$} \Comment{\texttt{\scriptsize merge theories}}
      \If{\Call{bic}{$E_{i + j}$} $\leq$ \Call{bic}{$E_i \cup E_j$}} \Comment{\texttt{\scriptsize verify improvement}}
        \State \textit{merged} $\gets$ \True
        \Break
      \EndIf
    \EndWhile

    \If{\textit{merged}}      \Comment{\texttt{\scriptsize{merge occurred}}}
      \State $\mathbb{E}$ $\gets$ \Call{update}{$E_i$, $E_j$, $E_{i + j}$}   \Comment{\texttt{\scriptsize update hierarchy}}
    \EndIf
  
  \Until{$\mid$ E $\mid > 1 \text{ } \wedge$ \textit{merged}}
  \Comment{\texttt{\scriptsize until the merge is successful}}
  \State $E \gets \Call{filter}{E, \alpha}$ \Comment{\texttt{\scriptsize{Filter final theory}}}
  \State $\Return \text{ } E$
\end{algorithmic}
\caption{\textsc{GLocalX}$(\mathbb{E}, \alpha)$}
\label{alg:ethica}
\end{algorithm}

\ethica\ is detailed in Algorithm~\ref{alg:ethica}.
Given a set of explanation theories $E$, a pairwise similarity function and a quality criterion, we sort logical theories by similarity in a queue $\mathbb{Q}$\footnote{$\mathbb{Q}$ is the set of sorted candidates theories.} (line 3).
Then, we sample a batch of data to merge the candidate theories (line 5).
Using batches instead of the whole training dataset favors diverse merges, as the merge procedure has different behaviors according to the data at hand.
In the merge loop, we pop the queue to find the most similar pair of theories whose merge satisfies a quality criterion (line 7) and, we run the merge operation (line 8), and if the merge is advantageous (line 9), the merged theory is kept\footnote{$E_{i + j}$ indicates the merged theories and does not refer to the sum of the indexes.}.
As a quality criterion we have selected the Bayesian Information Criterion (BIC)~\cite{wit2012all}, as it rewards models for their simplicity and accuracy.
BIC has been successfully adopted in various techniques, i.e., clustering, adopting bisecting hierarchical refinement of the model \cite{pelleg2000x,guidotti2015tosca}.
After a successful merge we replace the two mergees $E_i, E_j$ with the merged theory $E_{i + j}$ (line 13).
If no advantageous merge is found, \ethica\ halts.
This process is iterated until no more merges are possible (line 14).
Finally, explanations with low fidelity are filtered out to reduce the output size (line 15): we use a parameter $\alpha$ to indicate this per-class trimming threshold.
Specifically, we select the top-$\alpha$ explanations by fidelity, $\lceil \alpha^{-1} \rceil$ for positive and negative class, respectively.
In addition to this, we introduce $\alpha_q$, a relative trimming criterion discarding rules with fidelities under the $\alpha_q^{th}$ fidelity percentile.
Trimming explanations with low fidelity allows us to retain only the best explanations to provide in output.

\subsection{Finding Similar Theories}
\label{sec:ethica_similarity}
Selecting pairs of theories to merge requires the definition of a pairwise similarity function on logical explanation theories (line $2$, Algorithm~\ref{alg:ethica}).
To this aim, we define the similarity of two theories $E_1, E_2$ as the \textit{Jaccard similarity}~\cite{tan2007introduction} of their \textit{coverage} on a given instance set $X$:
\begin{equation*}
similarity_X(E_i, E_j) = \frac{|coverage(E_i, X) \cap coverage(E_j, X) |}{|coverage(E_i, X) \cup coverage(E_j, X) |}.
\label{eq:theories_similarity}
\end{equation*}
An explanation $e = \langle r = P \rightarrow y \rangle$ \textit{covers} an instance $x$ if the premise $P$ of $r$ is satisfied by $x$.
We extend the notion of coverage to explanation theories by saying that an explanation theory $E$ \textit{covers} an instance $x$ if there is at least an explanation $e \in E$ covering $x$.
$coverage(E, X)$ returns the set of records in $X$ covered by the set of explanations in $E$, i.e., $coverage(E, X) = \{x \in X \mid \exists \; e \in E. \ e\ \text{covers } x\}$.
Conversely, $covered(x, E)$ returns the set of explanations in $E$ with coverage on $x$, i.e., $covered(x, E) = \{e \in E \mid x \in coverage(e, \{x\})\}$.
As for $coverage$, we extend this notion to sets of records by saying that a record $x$ is \textit{covered} by an explanation theory $E$ if there is at least an explanation $e \in E$ that \textit{cover}s $x$.

The larger is the shared coverage of $E_i$ and $E_j$ on $X$, the more similar the two logical explanation theories are.
Coverage similarity is a two-faceted similarity measure that captures \textit{(i)} the premise similarity and \textit{(ii)} the coverage similarity.
The former is straightforward: rules with similar premises will have similar coverage.
The latter balances the premise similarity to avoid that rules with similar \mbox{premises but low coverage sway the similarity score.}

\subsection{Merging Explanation Theories}
\label{sec:ethica_merge}
The merge function allows \ethica\ to \textit{generalize} a set of explanation theories while balancing fidelity and complexity through approximate logical entailment.
As detailed in the following, the merge involves two operators, \textit{join} and \textit{cut}, to simultaneously generalize and preserve a high level of fidelity.
In particular, in the logical domain, generalization seldom involves premises relaxation or outright removal~\cite{DBLP:journals/ml/Furnkranz97}.
Thus, \ethica\ advances the state-of-the-art in exploiting also this kind of generalization.
We highlight that generalization comes at a fidelity cost, as the more general a set of premises is, the more likely it is to capture unwanted instances.
Pushing for generalization may pull down fidelity.
These contrasting behaviors are the focal point in a \ltg\ setting and must be dealt with accordingly.
We tackle this double-faced problem with a merge function that handles both rule generalization and fidelity.

We illustrate our proposal with a trivial example.
Suppose we have two explanation theories, $E_1 = \{e_1, e_2\}$ and $E_2 = \{e'_1, e'_2\}$, with $e_1, e'_1$ explaining a record $x_1$ and $e_2, e'_2$ explaining a record $x_2$.
\begin{align*}
  E_1 = \{ & e_1 = \{\text{age }{\geq} 25, \text{job = unemployed}, \text{amount }\leq{ 10k}\} {\rightarrow} \text{ deny}\; \\
    & e_2 = \{\text{age }{\geq} 50, \text{job = office clerk}\} {\rightarrow} \text{ deny} \} \\
  E_2 = \{ & e'_1 = \{\text{age }{\geq} 20, \text{job = manager}, \text{amount }>{8k}\} {\rightarrow} \text{ accept}\; \\
    & e'_2 = \{\text{age }{\geq} 40, \text{job = office clerk}, \text{amount }>{5k}\} {\rightarrow} \text{ deny} \}
\end{align*}
The resulting merge yields the following rules $E_{1+2}$:
\begin{align*}
  E_{1+2} =\{
    & e''_1 = \{\text{age }{\geq } 25, \text{job = unemployed}, \text{amount }\leq{ 10k}\} {\rightarrow} \text{ deny} \\
    & e''_2 = \{\text{age }{\in } \text{[20, 25]}, \text{job = manager}, \text{amount }{\in}{ [8k, 10k]}\} {\rightarrow} \text{ accept} \\
    & e''_3 = \{\text{age }{\geq } 40\} {\rightarrow} \text{ deny} \}
\end{align*}
In $E_{1+2}$, rules with equal predictions from different explanation theories have been generalized by relaxing their premises (rule $e''_3$) while rules with different predictions have been specialized by further constraining them (rule $e''_2$).
Specifically, $e''_1, e''_2$ results from a cut on $e_1, e'_1$, that is $e_1 - e'_1$.

\begin{algorithm}[t]
  \begin{algorithmic}[1]
    \Input $E_i, E_j$ explanation theories, $X$ batch 
    \Output $E_{(i + j)}$ explanation theory 

    \State E $\gets E_i \cup E_j$

    \For{$x \in X$}
        \State $C_i \gets$ \Call{covered}{$x$, $E_j$} \Comment{\texttt{\scriptsize retrieve rules in $E_i$ covering $x$}}
        \State $C_j \gets$ \Call{covered}{$x$, $E_j$} \Comment{\texttt{\scriptsize retrieve rules in $E_j$ covering $x$}}
        \State $C_= \gets$ \Call{non-conflicting}{$x$, $C_i$, $C_j$} \Comment{\texttt{\scriptsize non-conflicting rules in $C_i, C_j$ and covering $x$}}
        \State $C_{\neq} \gets$ \Call{conflicting}{$x$, $C_i$, $C_j$} \Comment{\texttt{\scriptsize non-conflicting rules in $C_i, C_j$ covering $x$}}
        
        \State E $\gets$ E $\setminus (E^i \cup E^j)$ 
        
        \State $E_= \gets$ \Call{\join}{$C_=$}
        \State $E_{\neq} \gets$ \Call{\cut}{$C_{\neq}$, $X$}
    
        \State E $\gets$ E $\cup$ $E_=$ $\cup$ $E_{\neq}$
    \EndFor
  \State $\Return \text{ } E$
\end{algorithmic}
\caption{\textsc{merge}$(E_i, E_j, X)$}
\label{alg:ethica_merge}
\end{algorithm}

More formally, given two explanation theories $E_i, E_j$, the \textit{merge} function applies two  operators on the explanation theories to derive a new theory approximately entailed by the two: the \join\ and the \cut\ operators.
The former allows merging \textit{non-conflicting} rules while the latter allows merging \textit{conflicting} rules~\cite{schmidhuber2015deep}.
A set of explanations $E = \{e_1, \dots, e_n\}$ part of a logical explanation theory is considered \textit{conflicting} on an instance $x$ if two or more of them cover an instance $x$ but lead to two different outcomes.
The merge of two logical explanation theories $E_i$ and $E_j$ applies the \join\ operator on \textit{non-conflicting} explanations and the \cut\ operator on \textit{conflicting} explanations iteratively on each instance in the batch.
We detail this process in Algorithm~\ref{alg:ethica_merge}.
The resulting set of explanations composes the new explanation theory $E_{i + j}$ (line 8, Algorithm~\ref{alg:ethica}).
The candidate merge is then tested for considering the equilibrium between fidelity and complexity with the Bayesian Information Criterion (BIC)~\cite{wit2012all} computation.
In our case, the model log-likelihood is computed as the rules fidelity, and the model complexity as the average rule length.

In the following we provide details of the \join\ and \cut\ operators.
The two operators move in opposite directions: \join\ generalizes explanations, possibly at a fidelity cost, while the \cut\ specializes explanations, possibly at a generalization cost.
In other words, the former allows generalization while the latter regularizes it.
Inspired by~\cite{ruggieri2013learning}, we define the \join\ and the \cut\ operators through an alternative representation of decision rules.
Let $\widehat{\mathcal{X}}_i$ be the set of subspaces on the feature $i$.
Given a decision rule $r = P \rightarrow y$ we have that any $P_i \in \widehat{\mathcal{X}}_i$ is a subspace on the feature $i$.
The premise $P$ of the rule identifies a \emph{\qp} \mbox{defined as a subspace of $\widehat{\mathcal{X}}^{(m)}$:}
\begin{equation*}
P = \{P_i, \dots, P_j\} \in \mathcal{P}(\widehat{\mathcal{X}}_1 \times \dots \times \widehat{\mathcal{X}}_m)
\label{eq:quasi_polyhedron}
\end{equation*}
We say that an instance $x$ \emph{satisfies} $P$ if $\forall P_i \in P.\ x_i \in P_i$, thus, $x$ satisfies $P$ if it lies in the subspace defined by the \qp.
We define the operators \emph{join} ($\oplus$) and \emph{cut} ($\ominus$) exploited by the \textit{merge} function based on this \qp\ interpretation.

\paragraph{Reasoning in the polyhedral space
The polyhedral interpretation lends itself to a straightforward approximate inference algorithm, which we call ``of inclusion''.
Given the equivalence of decision rules and quasi-polyhedra, a rule $s$ with quasi-polyhedron $P_s$ is inferred by another rule $r$ with quasi-polyhedron $P_r$ ($r \xrightarrow{} s$) if and only if its premises (and thus the record satisfying them) are implied by the other.
Simply put, all instances satisfied by $s$ are also satisfied by $r$.
Quasi-polyhedra-wise, it follows that $P_s \subseteq P_r$.
Hence, inference by inclusion produces flat reasoning paths in which the implied local rules are elevated to global explanations, yielding a ``global'' model just as local as before.
With this consideration in mind, we reject exact inference in favor of approximate inference in which \emph{join} and \emph{cut} perform approximate rule entailment.
}
      
\paragraph{\textbf{Join}} The \join\ operator aims to generalize a set of non-conflicting explanations relaxing their premises, hence generalizing the associated rules.
Given two quasi-polyhedra $P$ and $Q$, the \join\ ($\oplus$) is defined as follows:
\begin{equation*}
P \oplus Q = \{P_1 + Q_1, \dots, P_m + Q_m\}
\label{eq:join}
\end{equation*}
where:
%
%
%
\begin{equation*}
\footnotesize{
  P_i + Q_i = 
  \begin{cases}
    P_i \cup Q_i & \text{ non-empty intersection}\\
    [\min\{P_i \cup Q_i\}, \max\{P_i \cup Q_i\} ) & \text{ empty intersection}\\
    \emptyset & P_i \text{ is empty} \vee Q_i \text{ empty}\\
  \end{cases}
}
\end{equation*}

\noindent \textit{Example.} Consider the following two explanations:
\begin{align*}
  e_1 &= \{\text{age} \geq 50, \text{job = office clerk}\} \rightarrow \text{ deny} \\
  e_2 &= \{\text{age} \geq 40\} \rightarrow \text{ deny} \\
\end{align*}
The merge function applies the \join\ operator $e_1 \oplus e_2$ and returns
\begin{align*}
  e'_1 &= \{\text{age} \geq 40\} \rightarrow \text{ deny}\\
\end{align*}
The shared feature \textit{age} has a \textit{non-empty intersection} ($\text{age} \geq 50 \cap \text{age} \geq 40 \neq \emptyset$), hence it is generalized to encompass both premises according to the first case.
On the non shared feature $\text{job = office clerk}$ we have one empty quasi-polyhedron, hence it is removed according to the third case.

\paragraph{\textbf{Cut}} The \cut\ operator acts in a complementary fashion by slicing quasi-polyhedra.
Here, the goal is to preserve the better rules, and confine the lesser ones to subspaces in which they have high fidelity.
In other words, the aim is to remove overlaps between rules by subtracting them.
This directly translates to subtracting quasi-polyhedra.
Formally, given two quasi-polyhedra $P$ and $Q$, the \cut\ ($\ominus$) is defined as:
\begin{equation*}
P \ominus Q = \{P_1 - Q_1, \dots, P_m - Q_m\}
\label{eq:qp_diff}
\end{equation*}
where:
\begin{equation*}
  P_i - Q_i = 
  \begin{cases}
    \{P_i, \emptyset \} & Q_i \text{ empty}\\
    \{P_i, Q_i \setminus Q_i \} & \text{ otherwise}
  \end{cases}
\end{equation*}
Note that, unlike the \join\ operator, the \cut\ operator is not symmetric, hence $P \ominus Q \neq Q \ominus P$.
With our goal of preserving high fidelity rules and restrict lower fidelity rules in mind, it is straightforward to select subtracted and subtracting polyhedron, with the two rules being the one with higher and lower fidelity, respectively.

\noindent \textit{Example.} Consider the following two explanations:
\begin{align*}
e_1 &= \{\textit{age} \geq 25, \textit{job = unemployed}, \textit{amount} \geq 10k\} \rightarrow \textit{deny} \\
e_2 &= \{\textit{age} \geq 20, \textit{job = manager}, \textit{amount} > 8k\} \rightarrow \textit{accept}
\end{align*}
where the first one is the dominant one, that is, the one with highest fidelity.
The merge function applies the \cut\ operator $e_1 \ominus e_2$ and returns
\begin{align*}
  e_1 &= \{\textit{age} \geq 25, \textit{job = unemployed}, \textit{amount} \geq 10k\} \rightarrow \textit{deny} \\
  e'_1 &= \{\textit{age} \in \textit{[20, 25]}, \textit{job = manager}, \textit{amount} \in [8k, 10k]\} \rightarrow \textit{accept} \\
\end{align*}
We note that $e_1$ is preserved as-is while the premises of $e_2$ are further constrained to reduce overlap.
Namely, \textit{age} is constrained to remove overlap on \textit{age} $\geq$ 25 and \textit{amount} is constrained to remove overlap on \textit{amount} $\geq$ 10k.

\medskip
We highlight that $\oplus$ and $\ominus$ strictly operate on the premises of rules, ignoring their outcome, which is preserved after the merge.
Moreover, \textsc{join} and \textsc{cut} 
produce sets of different cardinality.
While \textsc{cut} preserves the number of rules, join indeed lessens it by subsuming it in a smaller more general explanation set.
    
\subsection{Interpretable Classification}
\label{sec:classification}
The global explanation theory $E$ mimes the black box decision logic for predicting a set of instances.
By construction, $E$ is not exhaustive nor mutually exclusive, i.e., not all records are necessarily covered by at least one rule while covered records may be covered by more than one rule.
We address the former case with a default majority rule and the latter with a voting schema.

We exploit $E$ as a rule-based classifier to replace the opaque black box $b$ in the prediction task.  
Moreover, $E$ may also be used for explaining the decision of the black box $b$ on a single instance $x$.
Clearly, as a global explanation, it can be less accurate than the local one, due to manipulations applied to capture the global behavior of the black box.        

It is important to highlight that, given an instance $x$ we could have more than one decision rule (explanation) in $E$ that satisfies $x$, and some of them can also be conflicting.
As a consequence, to apply $E$ for classifying $x$ or explaining $b(x)$ we need a mechanism for deriving a decision when multiple covering rules have different outcomes.
In line with the literature~\cite{yin2003cpar}, we employ the following: given a record $x$, we select the rule with the highest accuracy among the ones covering $x$, and use its outcome as prediction.
Note that this mechanism replicates the Laplacian scoring schema proposed in~\cite{yin2003cpar} and the \emph{Falling Rule List} schema introduced by~\cite{lakkaraju2016interpretable} setting a number of voting rules to one.

\begin{table}[t!]
\centering
\small
\begin{tabular}{cccccc}
\toprule
& instances & features & $|X_{\mathit{bb}}|$ & $|X_{\mathit{le}}|$ & $|X_{\mathit{ts}}|$ \\
\midrule
\adult  & 32,560  & 10 & 22,791 & 6,837 & 2,930 \\
\compas & 7,214   & 18 & 5,048  & 1,514 & 649   \\
\german & 1,000   & 19 & 699    & 209   & 89    \\
\diva   & 8,000   & 88 & 4800   & 1600  & 1600  \\
\bottomrule
\end{tabular}
\caption{Datasets statistics.}
\label{tbl:datasets}
\end{table}

\begin{figure}[t]
  \centering
  \includegraphics[width=0.75\linewidth]{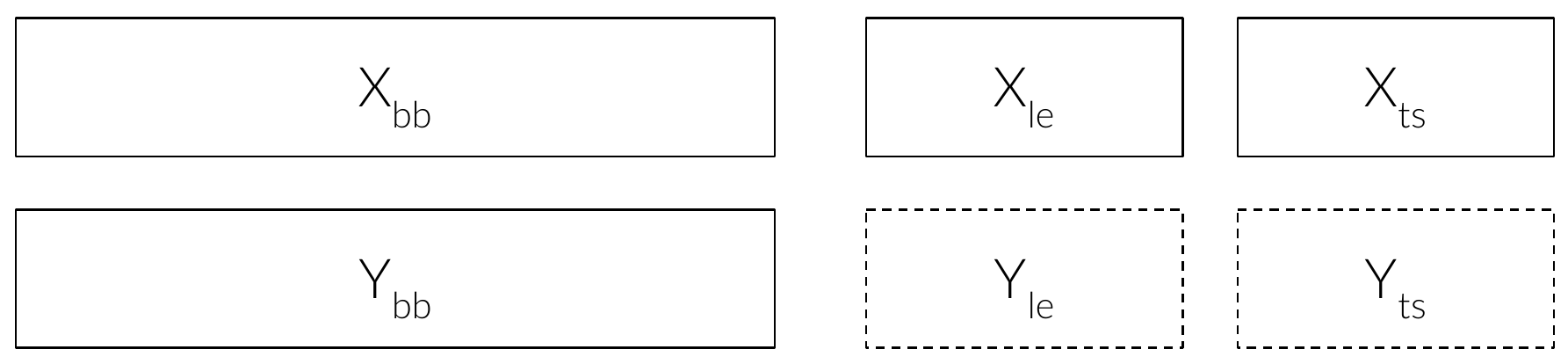}
  \caption{Validation schema: $X_{\mathit{bb}}$ is used for training the black box, $X_{\mathit{le}}$ is the partition to explain, while $X_{\mathit{ts}}$ is reserved for validating the fidelity and the accuracy. The dashed line indicates the labels predicted by the black box and on which the model fidelity is estimated.}
  \label{fig:validationschema}
\end{figure}

\section{Experiments}
\label{sec:experiments}
In this section, after presenting the experimental setup, we report an analysis of \ethica\ on a set of standard benchmark datasets and a real-world proprietary dataset, and compare \ethica\ to native baselines and state-of-the-art global explainers.
Moreover, we analyze \ethica\ in two settings: in the former, we provide \ethica\ with a restricted number of rules while in the latter, we do not provide \ethica\ with any data.
\ethica\ has been developed in Python\footnote{The Python implementation is available at \href{https://github.com/msetzu/glocalx}{github.com/msetzu/glocalx}}. 
The experiments were performed on \code{16-core Intel(R) Xeon(R) CPU E5-2620 v4 @ 2.10GHz}, \code{128 GB} of RAM.
For the local rule extraction we selected \lore~\cite{guidotti2019factual} due to its high fidelity and stability~\cite{guidotti2019stability}. 
Alternatives are \anchors~\cite{ribeiro2018anchors} or generating a random local neighborhood like in \lime~\cite{ribeiro2016should} and the using locally a rule based classifier.

\subsection{Experimental Setup}
\label{sec:expsetup}
We showcase the proposed approach on three benchmark datasets frequently used in the literature, namely \adult\footnote{\href{https://archive.ics.uci.edu/ml/datasets/Adult}{https://archive.ics.uci.edu/ml/datasets/Adult}}, \compas\footnote{\href{https://github.com/propublica/compas-analysis}{https://github.com/propublica/compas-analysis}}, and \german\footnote{https://archive.ics.uci.edu/ml/datasets/statlog+(german+credit+data)}, and a real world proprietary dataset, \diva\footnote{\href{https://kdd.isti.cnr.it/project/diva}{https://kdd.isti.cnr.it/project/diva}}.
The \adult\ dataset includes $48,842$ instances with information like age, job, marital status, race, capital loss, capital gain, etc.
The labels have values ``$<$=50K'' or ``$>$50K'', indicating whether the person will earn more or less than $50k\$$ in this fiscal year.
The \compas\ recidivism dataset contains the features used by the COMPAS algorithm for scoring defendants and their risk of recidivism (Low, Medium and High), for over $4,000$ individuals.
We have considered the two classes ``Low-Medium'' and ``Medium'' as equivalent to map the task into a binary classification task.
The dataset includes features like: age, sex, race, priors count, days before screening arrest, length of sentence, charge degree, etc.
In \german\ each one of the $1,000$ persons is classified as a ``good'' or ``bad'' creditor according to attributes like age, sex, checking account, credit amount, loan purpose, etc.
Finally, \diva\ is a privately released dataset on fraud evasion, periodically issued by the Italian Ministry of Economics\footnote{Due to privacy and legal concerns, we are not allowed to publicly release the dataset.}.
\diva\ records financial activities for more than $12,000$ citizens, including their past financial credit score, declared income and property value, debt and several taxation detailed infos.
The labels mark fraudulent citizens.
These datasets contain both categorical and continuous features.
Missing values, if present, were replaced by the mean for continuous features and by the mode for categorical ones.
Details are reported in Table~\ref{tbl:datasets}.
Similarly to the training/test classical split, we split the dataset into three partitions: $X_{\mathit{bb}}$, the set of records to train a black box model; \xte, the set of locally explained records which is also used as reference set by \ethica\ for calculating coverage and fidelity in training phase, and by the global competitors for training; $X_{\mathit{ts}}$, the held-out set of records to validate the fidelity and accuracy of \ethica.
\mbox{Figure~\ref{fig:validationschema} depicts this three-way split.}

We validate \ethica\ on its ability to mime Deep Neural Networks (DNN), Random Forests (RF) and Support Vector Machines (SVM).
The black boxes have been trained on $X_{\mathit{bb}}$ with grid searches on a $3$-fold cross-validation schema.
Given the poor performances of DNNs and SVMs on \texttt{diva} and \texttt{german} datasets, they were experimented upon only on Random Forests. 
Performance in terms of accuracy of these black boxes are reported in Table~\ref{tbl:black_VS_crystal} ($3^{rd}$ column).
Given the novelty of the problem and the lack of local to global explainer in the literature\footnote{\lime~\cite{ribeiro2016should} and some other recent works extending \lime\ claim to obtain a global explanation by joining local feature importance but in fact the resulting model is just a set of numbers and cannot be used to replace the black box, nor expresses in logical form the logic adopted by the black box for taking decisions.}, we compare \ethica\ against natively global frameworks: Decision Tree~\cite{quinlan1993c4} (\textsc{dt}), Pruned Decision Tree\footnote{A pruned decision tree is a decision tree with maximum dept equals to four. We adopt four as maximum dept as it is the measure used in Optimal Decision trees~\cite{bertsimas2017optimal}. We do not compare with Optimal Decision trees due to the complexity of running the models that requires a particular architecture and the non public availability of the code. } (\textsc{pdt}) in a \quinlan~\cite{quinlan1993c4} implementation, and \cpar~\cite{yin2003cpar}, a rule-based classifier (\textsc{rb}) implemented in the \textit{LUCS-KDD} library\footnote{\href{https://cgi.csc.liv.ac.uk/~frans/KDD/Software}{https://cgi.csc.liv.ac.uk/$\sim$frans/KDD/Software}}.

In addition, we also experiment with \ethica\ in a particular setting in which only local explanations $e_1, \ldots, e_n$ are provided, and no data $X_{le}$ is available.
We call this setting \emph{synthetic} and name it \ethica* for short.
In this case, we assume to have some information on the feature distribution, and employ a data generation and sampling schema similar to~\cite{craven1996extracting} to construct a training dataset for \ethica\ of the same size of \xte.
Specifically, the joint distribution has been estimated with a Gaussian density estimator, which has then been randomly sampled to build a training set for \ethica*.
This dataset is then provided as input, and \ethica\ is executed as previously defined.
With \ethica*, we wish to grasp the learning abilities of \ethica\ when a minimal input comprised exclusively of the local rules and data distributions is provided.
A minimal assumption of knowledge on the data distribution is needed to validate the merging procedure.
Relaxing this assumption would reduce the merging problem in a logical inference one, as no quantitative evaluation of any merge can be performed, leaving only intra-rule implication as a form of rule generalization.
In other words, when no knowledge on the data is available, the \ltg\ problem is equivalent to a logic inference problem.

\begin{table}[t]
\centering
\footnotesize
\begin{tabular}{cc|c|ccc}
  \toprule
  \textbf{Dataset} & \textbf{Black Box} & \textbf{Accuracy} & \textbf{\textsc{dt} Accuracy} & \textbf{\textsc{dt} Size} & \textbf{\textsc{dt} Length} \\
  \midrule
  \multirow{3}{*}{\adult}   & DNN & 0.868 & & & \\
                            & RF  & 0.860 & 0.813 & 5452   & $16.611 \pm 6.150$  \\
                            & SVM & 0.860 & & &   \\
  \midrule
  \multirow{3}{*}{\compas}  & DNN & 0.611 & & &  \\
                            & RF  & 0.548 & 0.587 & 1514   & $6.271 \pm 1.985$  \\
                            & SVM & 0.557 & & & \\
  \midrule
  \multirow{1}{*}{\german}  & RF  & 0.753 & 0.896 & 68    & $5.294 \pm 2.065$  \\
  \midrule
  \multirow{1}{*}{\diva}    & RF  & 0.900 & 0.848 & 934   & $11.233  \pm 5.268$  \\
  \bottomrule
\end{tabular}
\caption{Black box accuracy on the various dataset ($3^{rd}$ column), and performance of a decision tree (\textsc{dt}) trained on  
the black box models and \textsc{dt} are trained on $X_{\mathit{bb}}, y_{bb}$ where $y_{bb}$ is the ground truth.
Both the black box models and the decision tree are evaluated on the $X_{\mathit{ts}}$.}
\label{tbl:black_VS_crystal}
\end{table}

\subsection{Evaluation Measures}
Given a black box $b$ and the explanation theory $E$ returned either by \ethica\ or by a global transparent model, we consider the following properties in evaluating its performance:
\begin{itemize}
  \item $\mathit{fidelity}(Y, \widehat{Y}) \in [0, 1]$ where $Y$ and $\widehat{Y}$ are the predictions returned by the rules in $E$ or by the black box $b$, respectively.
  The fidelity is the accuracy of the transparent global model in approximating the behavior of $b$~\cite{freitas2014comprehensible,lakkaraju2016interpretable}.
  \item $\mathit{size}(E) = | E |$ is the \textit{number of explanations} in the explanation theory $E$.
  \item $\mathit{length}(E) \in \mathbb{R}^+$ is the \textit{mean number of premises} of the rules in $E$.
  \item $\mathit{accuracy}(Y, Y^*) \in [0, 1]$ where $Y$ is the classification returned by the rules in $E$ (or by the black box $b$), and $Y^*$ are the real labels. It answers the questions: how good is the transparent model represented by $E$ in solving the classification problem? Can we replace $b$ with $E$?
\end{itemize}

If not differently specified, the results in the rest of this section use $X_{bb}$ to train the black box classifiers, $X_{le}$ to extract local explanations\footnote{Specifically, one explanation per record: the number of local explanations is directly inferred by the size of $X_{le}$ in Table~\ref{tbl:datasets}.}, run \ethica\ and learn the global interpretable models \textsc{dt}, \textsc{pdt} and \textsc{rb}, and $X_{ts}$ for evaluating the fidelity.
%
Thus, the \textit{fidelity} and the \textit{accuracy} refer to $X_{ts}$, while the \textit{size} and \textit{length} refer to models learned on $X_{le}$.
Experiments have been run with a batch size of $128$.\footnote{Smaller batches are not large enough for  performing reasonably accurate inference, while larger sets tend to reduce the diversity and number of merges.
In a preliminary stage of the research we have tested several batch sizes in $\{2^i\}, i \in \{1, \dots, 8\}$, with $128$ yielding satisfying results on all datasets.}

\subsection{Empirical Motivation of the Local to Global Explanation Problem}
\label{sec:motivation}
Besides the motivation presented in the introduction and in the problem definition section we show here an empirical reason for the local to global explanation problem.
Table~\ref{tbl:black_VS_crystal} shows the performance of an interpretable model, i.e., a Decision Tree (\textsc{dt}), trained on the same training set of the black box, i.e., $X_{\mathit{bb}}, Y_{bb}$ where $Y_{bb}$ are the real labels.
The results highlight that although \textsc{dt} has a high accuracy ($4^{th}$ column), it is usually lower than the accuracy of the black box classifiers ($3^{rd}$ column) in most of the cases (justifying the black box usage).
At the same time, despite being interpretable, the \textsc{dt} does not guarantee a real explicability.
Indeed, the good levels of accuracy come at the cost of a very high \textit{complexity} ($5^{th}, 6^{th}$ columns) of the learned model.
This is testified by the high \textit{size} (number of rules derived from the \textsc{dt}) and average rule length of the \textsc{dt}, making the overall transparent model non-interpretable in practice. 
We show in the following how \ethica, that works on a much smaller portion of data, is able to reach comparable performance in terms of fidelity while maintaining an admissible complexity of the explanation theory returned.

\begin{table}[t!]
  \centering
  \small
  \begin{tabular}{@{}lll@{}}
    \toprule
    \textbf{Feature} & \textbf{Recidivous} & \textbf{Non-recidivous} \\
    \midrule
    prior offenses           & \textbf{9.89} $\pm$ \textbf{5.18}  & $2.75 \pm 4.23$ \\
    past recidivous (\%)     & \textbf{74} $\pm$ \textbf{0.43}    & $0.42 \pm 0.49$ \\
    violent recidivous (\%)  & \textbf{19} $\pm$ \textbf{0.39}    & $0.09 \pm 0.29$ \\
    african-american (\%)    & \textbf{70} $\pm$ \textbf{0.45}    & $0.50 \pm 0.50$ \\
    caucasian (\%)           & $21 \pm 0.41$                      & \textbf{33} $\pm$ \textbf{0.47}\\
    \bottomrule
  \end{tabular}
  \caption{Sensitive features for the recidivous and non-recidivous group.
  Average number of \texttt{prior offenses} and percentage of past recidivists, violent recidivists, african-americans and caucasians.}
  \label{tbl:qualitative}
\end{table}

\subsection{Qualitative Evaluation}
In this section, we show an example of explanation theory $E$ yielded by \ethica.
We use as example the explanation theory returned by \ethica, for explaining the behavior of the RF black box on \compas, $\alpha = 6$.
\ethica\ achieves a fidelity of $0.86$ with a set of $6$ rules inferred from $1,515$ starting local rules:
\begin{align*}
  {E} = \{
      e_1 = \{&\textit{prior offenses} \geq 4, \textit{age} < 45, \\
            & \textit{sex = male}\} \rightarrow \textit{ Non-recidivous} ;\\
      e_2 = \{& \textit{prior offenses} \geq 4, \textit{age} < 45, \textit{sex = male}, \\
            & \textit{charge degree} = light\} \rightarrow \textit{Non-recidivous}; \\
      e_3 = \{&\textit{age} \leq 43, \textit{prior offenses } < 4, \textit{past recidivous} = no,\\
            & \textit{sex = male},
            \textit{charge degree} = light\} \rightarrow \textit{ Non-recidivous} ;\\
      e_4 = \{&\textit{past time in prison } < 8\} \rightarrow \textit{Recidivous}; \\
      e_5 = \{&\textit{priors count } < 4,  \textit{past time in prison } < 1\} \rightarrow \textit{Recidivous};\\
      e_6 = \{&\textit{priors count } < 2,  \textit{past time in prison } < 8\} \rightarrow \textit{Recidivous};\\
      \}&
\end{align*}

The explanations for the two classes show very different behaviors.
For the \texttt{Non-recidivous} class, explanations are rather lengthy, and account for the defendant's prior offenses, age, current charge and prior recidivous behavior.
Young men with a handful of light crimes and previous non-recidivous behavior appear to be the demographic of the non-recidivous behavior.
The \texttt{Recidivous} class shows different explanations, with brief rules involving exclusively the previous time in prison and the previous offenses.

On first sight, these explanations appear to be more coarse and do not show any self-evident bias towards sensitive features, e.g., race, of the defendants.
Table~\ref{tbl:qualitative} reports average values for the records covered by the two sets of explanations.
The recidivous explanations target, as expected, defendants with a high number of prior offenses ($9.89$ on average), high past recidivism ($74\%$) and past violent recidivism ($19\%$).
Intuitively, these are highly predictive features which one might expect to lead to future recidivism.
Moreover, the explanations indicate a possible bias against african-americans, which appear to be recidivous at a much higher rate ($70\%$) than caucasians ($50\%$).

\begin{table}[t!]
   \centering
   \footnotesize
   \setlength{\tabcolsep}{1mm}
   \begin{tabular}{cccccc|cccccc}
     \toprule
     & \textbf{$b$} & $\alpha_q$ & \textbf{fidelity} & \textbf{size} & \textbf{length} & & \textbf{black box} & $\alpha_q$ & \textbf{fidelity} & \textbf{size} & \textbf{length}\\
     \midrule
     \parbox[t]{2mm}{\multirow{14}{*}{\rotatebox[origin=c]{90}{\adult}}} & \multirow{4}{*}{DNN}         &   75 &  .872 &   53 &     $6.73 \pm 1.84$ &  \parbox[t]{2mm}{\multirow{14}{*}{\rotatebox[origin=c]{90}{\compas}}} & \multirow{4}{*}{DNN}  &   75 &  .769 &   16 &     $2.93 \pm 0.99$ \\
                                                                                                    &  &   90 &  .905 &   17 &     $6.43 \pm 1.76$ &  &  &   90 &  .751 &    8 &     $2.71 \pm 0.69$ \\
                                                                                                    &  &   95 &  .911 &   10 &     $5.33 \pm 1.15$ &  &  &   95 &  .757 &    5 &     $2.25 \pm 0.43$ \\
                                                                                                    &  &   99 &  .909 &    3 &     $7.00 \pm 0.00$ &  &  &   99 &  .759 &    4 &     $2.33 \pm 0.47$ \\
                                                                                                    &  & & & & &    &   & & & & \\[-1ex]
      & \multirow{4}{*}{RF}   &   75 &  .845 &   27 &    1$0.25 \pm 1.03$ &  & \multirow{4}{*}{RF}     &   75 &  .829 &   27 &     $5.57 \pm 1.52$ \\
                                                                                                    &  &   90 &  .862 &   11 &    1$0.80 \pm 0.60$  &  &  &   90 &  .873 &    9 &     $5.37 \pm 1.49$ \\
                                                                                                    &  &   95 &  .865 &    7 &    1$1.00 \pm 0.57$  &  &  &   95 &  .874 &    6 &     $5.20 \pm 1.72$ \\
                                                                                                    &  &   99 &  .867 &    2 &    1$2.00 \pm 0.00$  &  &  &   99 &  .875 &   10 &     $8.00 \pm 0.00$ \\
      &  & & & & &    &   & & & & \\[-1ex]
      & \multirow{4}{*}{SVM}  &   75 &  .859 &   56 &     $4.67 \pm 1.39$ &  & \multirow{4}{*}{SVM} &      75 &  .827 &   19 &     $6.22 \pm 1.54$ \\
                                                                                                    &  &   90 &  .868 &   22 &     $4.61 \pm 1.25$ &  &   &   90 &  .861 &    7 &     $5.66 \pm 0.47$ \\
                                                                                                    &  &   95 &  .902 &   10 &     $4.22 \pm 1.13$ &  &  &   95 &  .860 &    4 &     $5.66 \pm 0.47$ \\
                                                                                                    &  &   99 &  .865 &    3 &     $4.50 \pm 0.50$ &  &  &   99 &  .851 &    2 &     $6.00 \pm 0.00$ \\ \midrule
       \parbox[t]{2mm}{\multirow{4}{*}{\rotatebox[origin=c]{90}{\diva}}}     & \multirow{4}{*}{RF}   &     75 &  .855 &   48 &     $3.91 \pm 0.96$ & \parbox[t]{2mm}{\multirow{4}{*}{\rotatebox[origin=c]{90}{\german}}}  & \multirow{4}{*}{RF}    &   75 &  .786 &    2 &     $4.00 \pm 0.00$ \\
                                                                                                    &  &   90 &  .873 &   16 &     $3.86 \pm 0.71$ &  &  &   90 &  .786 &    2 &     $4.00 \pm 0.00$ \\
                                                                                                    &  &   95 &  .806 &    9 &     $3.75 \pm 0.82$ &  &  &   95 &  .786 &    2 &     $4.00 \pm 0.00$ \\
                                                                                                    &  &   99 &  .814 &    3 &     $4.00 \pm 1.00$ &  &  &   99 &  .786 &    2 &     $4.00 \pm 0.00$ \\
      
     \bottomrule
   \end{tabular}
  \caption{Impact of the filter parameter $\alpha_q$ on \ethica\ performance in terms of fidelity, size, and length and various datasets and black box classifiers. We observe that when considering simultaneously the three evaluation measures the best performance are reached for $\alpha_q = 95$, suggesting that few rules are sufficient for mimicking the global black box behavior.}
   \label{tbl:alpha}
 \end{table}

\subsection{Impact of the Filter Parameter}
First of all, we report an analysis of the impact of the filter parameter used to filter out rules from the final explanation theory returned by \ethica.
Here we use the relative trimming criterion $\alpha_q$ instead of the absolute one to minimize dataset-specific dependencies and study the effects of the filter across the fidelity distribution.
Table~\ref{tbl:alpha} reports the \textit{fidelity} and complexity (\textit{size} and \textit{length}) of \ethica\ varying $\alpha_q$ for the various datasets and black box models analyzed.
The fidelity is lower for values of $\alpha_q$ around $75$ and peaks around $\alpha_q = 95$, suggesting that a large number of rules may mislead the predictions of the model.
We can attribute this behavior to the use of batches in the construction phase: as stated in Section~\ref{sec:ethica}, the highest-fidelity rule is selected by verifying the fidelity of the rules on the batch at hand.
Hence, poor rules with good performance only on single batches may leak into the final model.
With respect to the complexity we observe a consistent decrease in size and a slight increase in explanation length across datasets and black boxes.
Thus, more complex rules seem to compensate for less rules in the explanation theory.

\begin{figure}[t]
  \centering
  \includegraphics[width=0.5\linewidth]{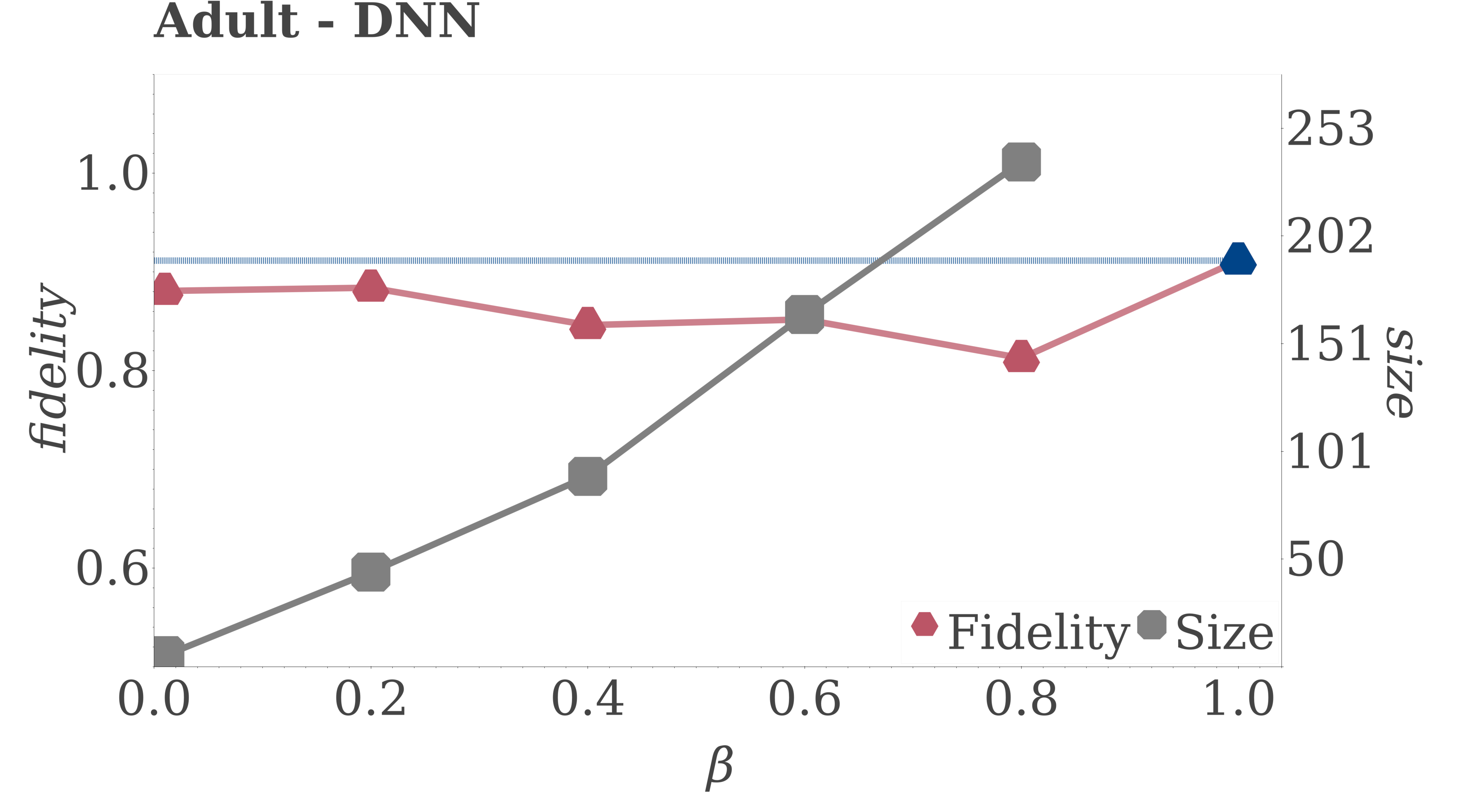}%
  \includegraphics[width=0.5\linewidth]{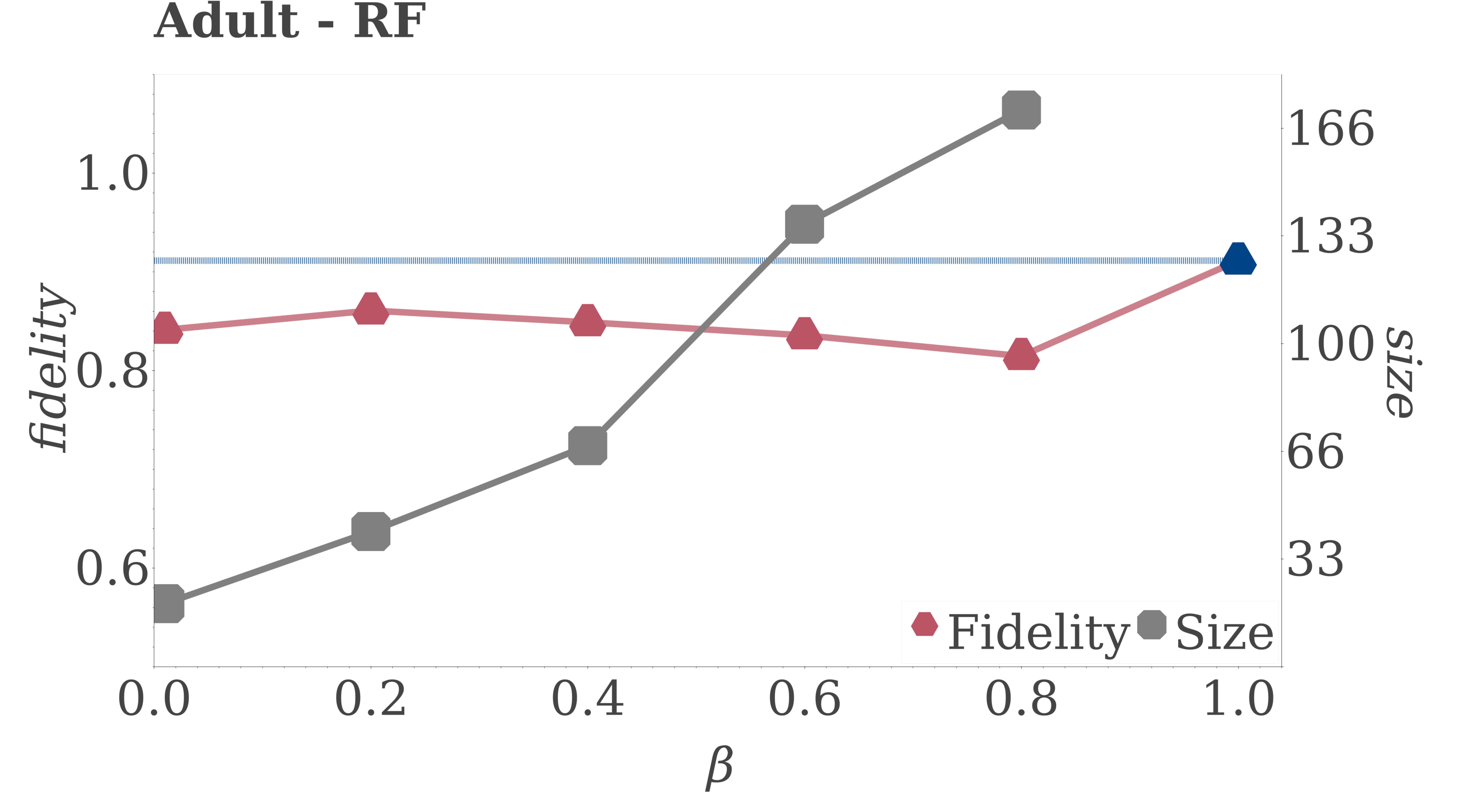}
  \includegraphics[width=0.5\linewidth]{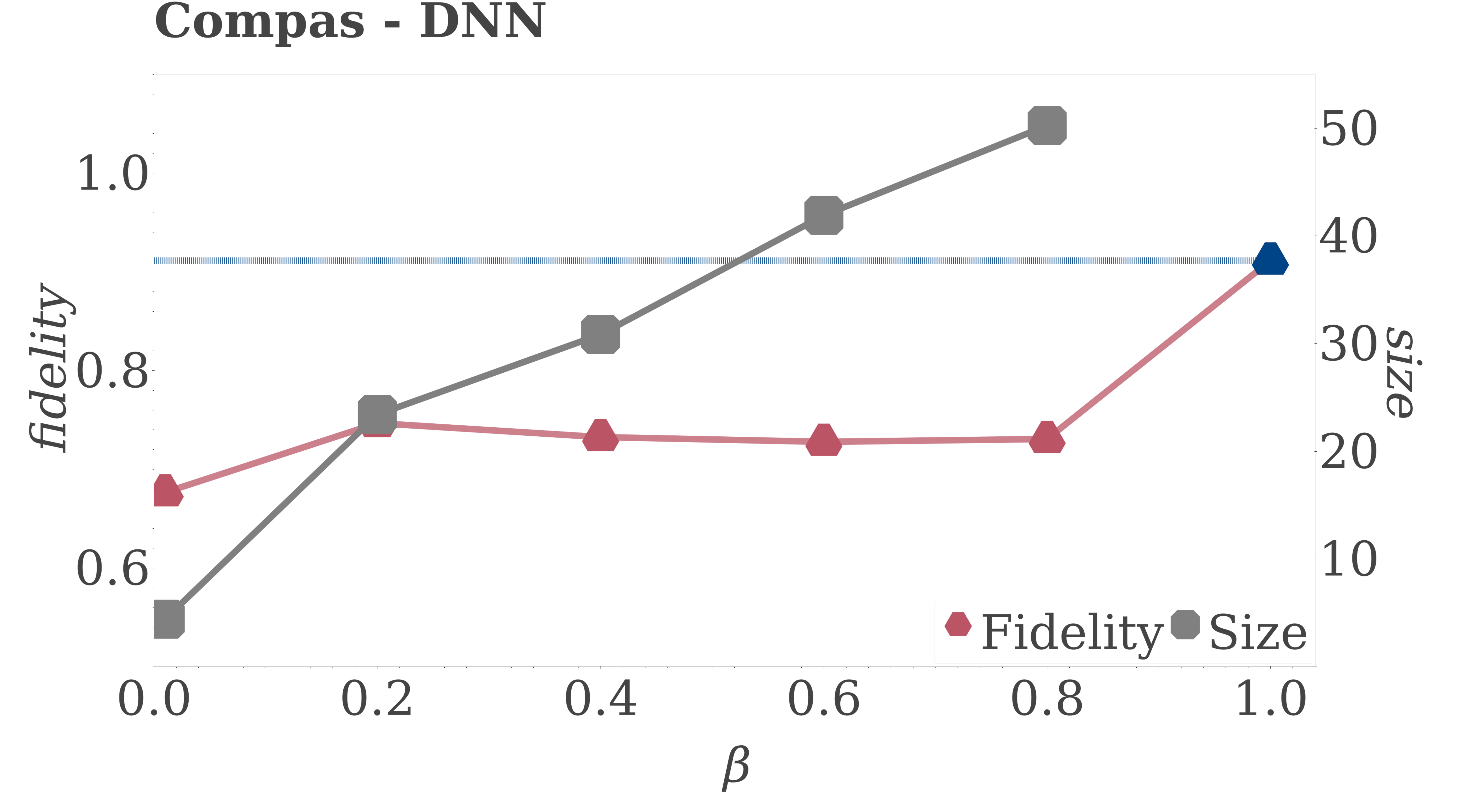}%
  \includegraphics[width=0.5\linewidth]{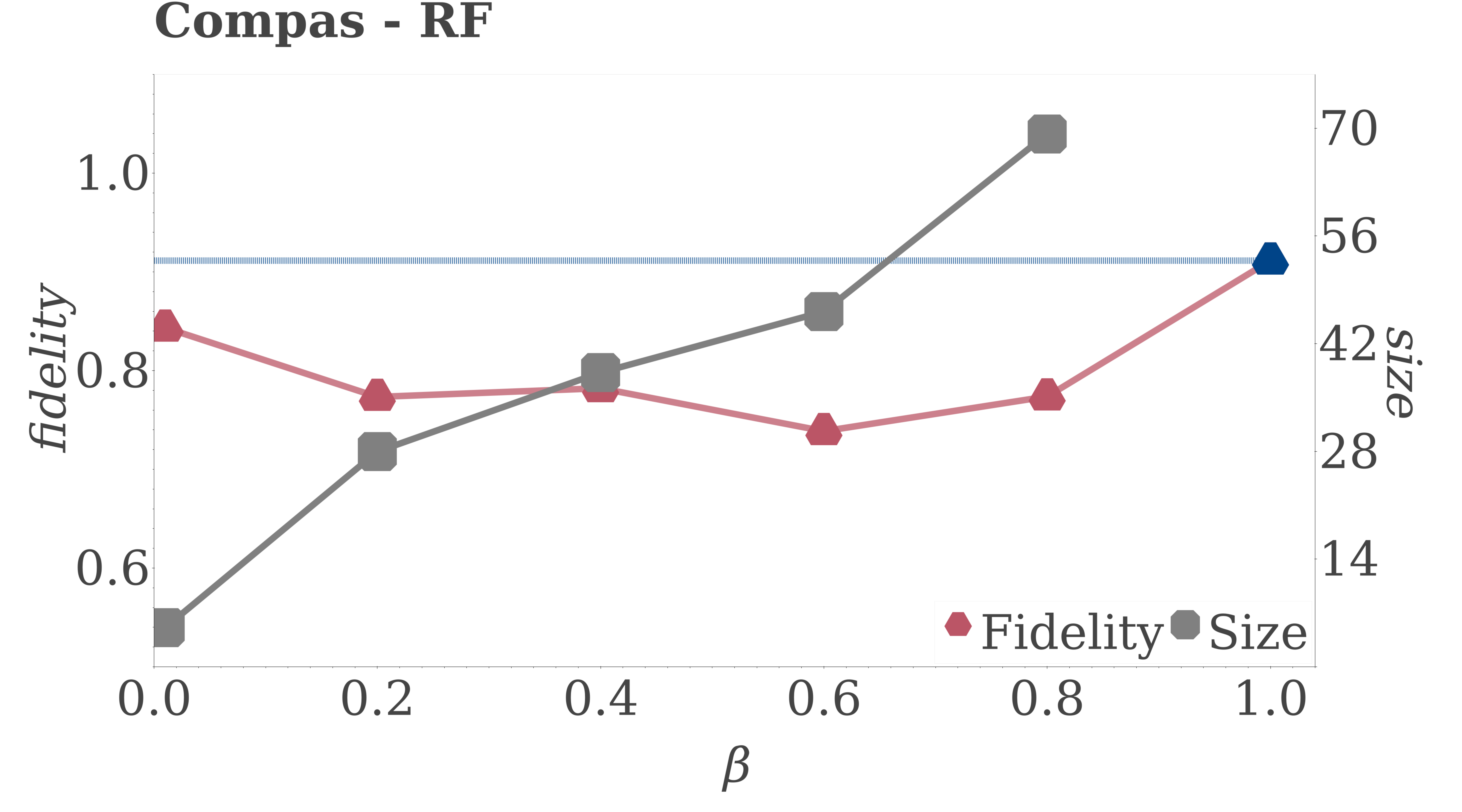}
  \includegraphics[width=0.5\linewidth]{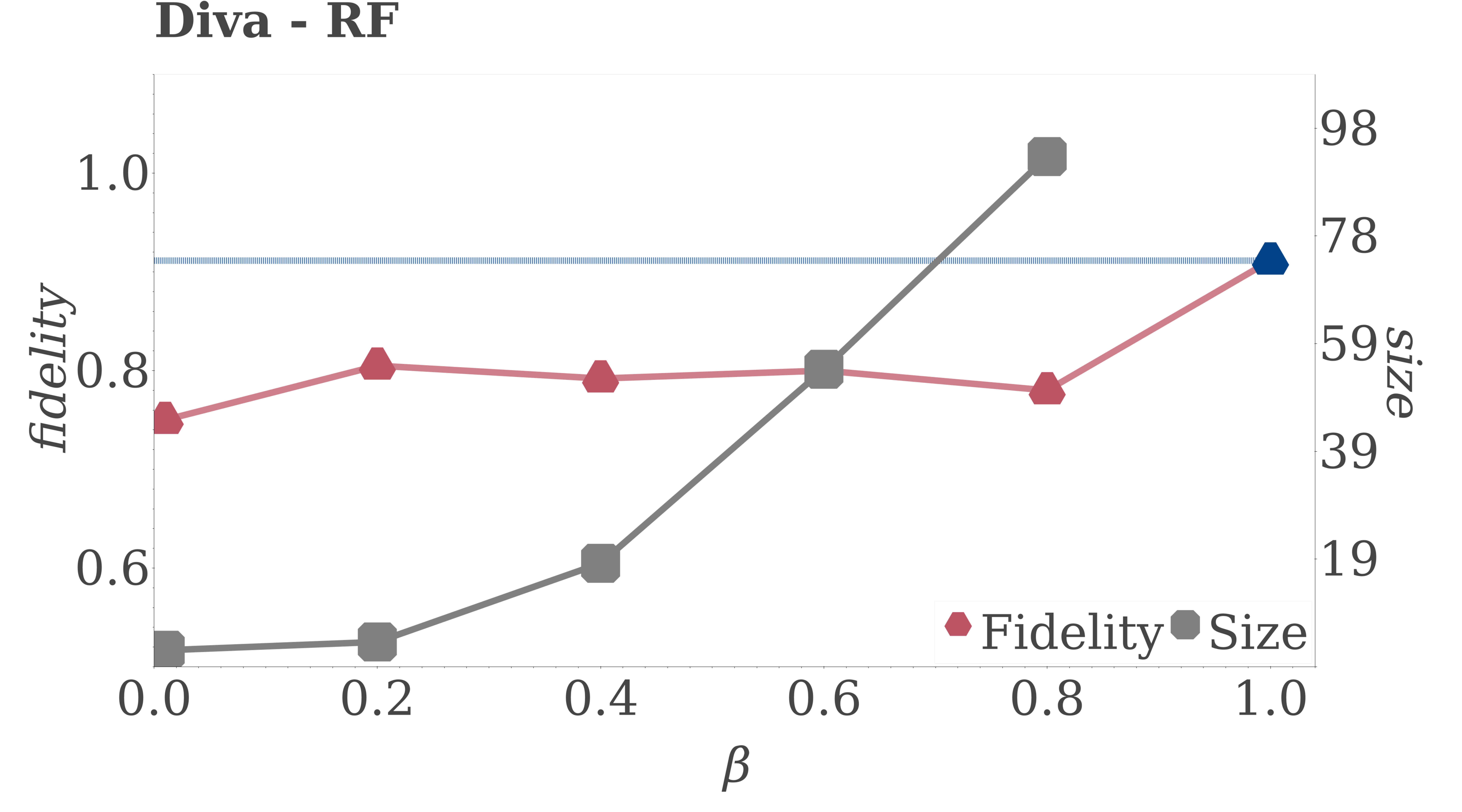}%
  \includegraphics[width=0.5\linewidth]{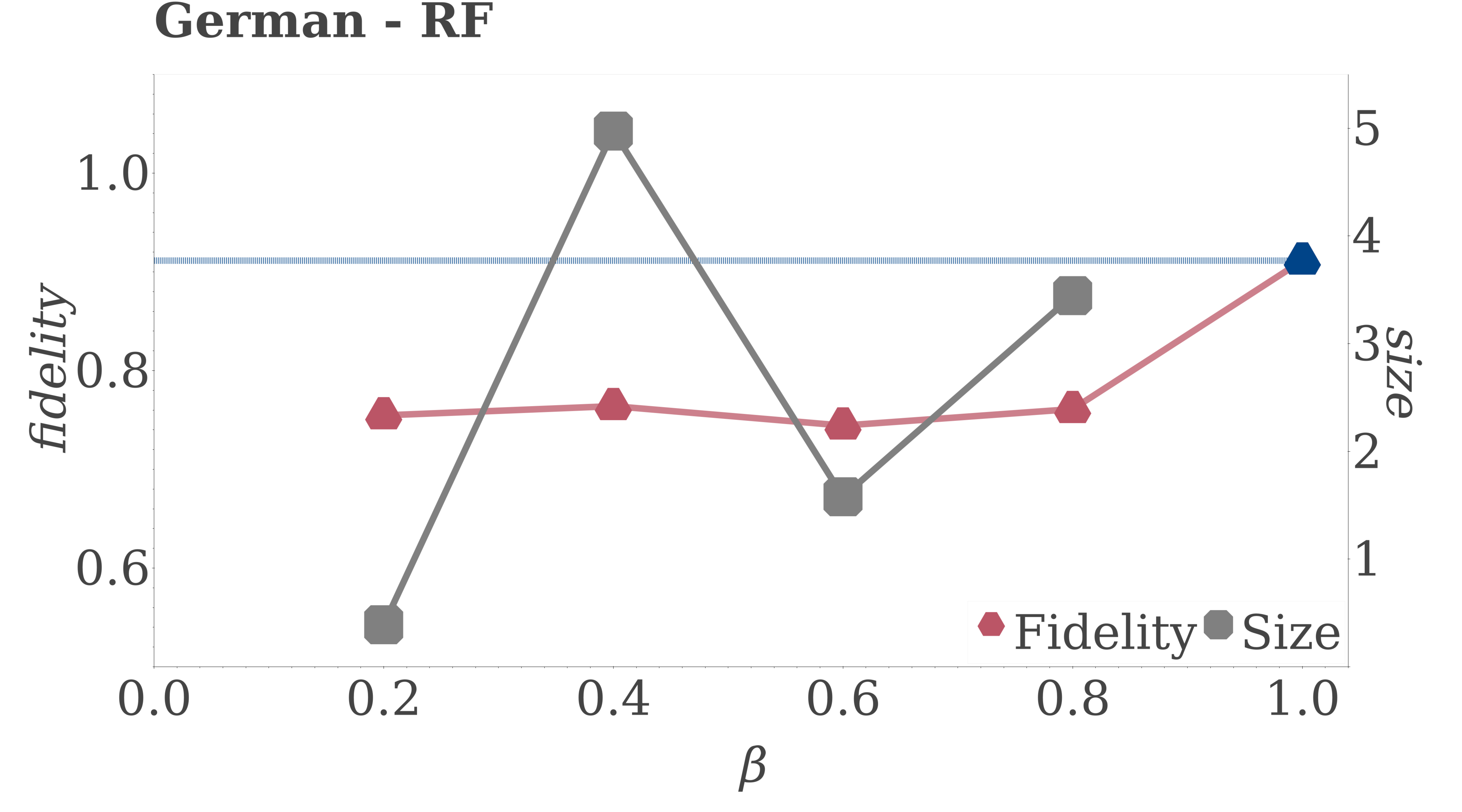}
  \caption{Effect of the number of local rules in \ethica\ with $\alpha_q = 50$.
  The red line denotes the fidelity, while the gray the size of \ethica varying the number of rules.
  The blue line denotes the fidelity of \ethica\ using all the available rules.}
  \label{fig:undersample}
\end{figure}

\subsection{Effect of the Number of Local Rules}
As second experiment we analyze the effect on the performance of \ethica\ of a reduced number or local explanation rules $e_1, \ldots, e_k$ ($k<n$) is provided in input with respect to considering a large set or all the available rules form $X_{le}$.
This scenario is of particular interest for applications where extracting rules is costly, there are additional constraints on the computation time, or simply there is a low number of available local explanations.
In practice, we provide as input to \ethica\ a subset of the rules available extracted with \lore\ from $X_{le}$.
Specifically, we provide \ethica\ with $\beta = \{1\%, 20\%, 40\%, 60\%, 80\%\}$ of the available rules, randomly sampling them in 10 independent trials\footnote{\german, which has a low number of rules, has been tested for $\beta = \{20\%, 40\%, 60\%, 80\%\}$}.
Figure~\ref{fig:undersample} reports fidelity and the size averaged over trials.
To lessen the impact of the $\alpha$-trimming, we report results with $\alpha = q^{50}\%$, i.e., instead of selecting the top-$\alpha$ explanations, we trim those with fidelity under the $50^{th}$ percentile.
\ethica\ shows small fluctuations in fidelity (left axis).
In particular, varying $\beta$ we observe an overall slight fidelity improvement.
While there is a small drop in fidelity, these results suggest that \ethica\ can be used with smaller sets of rules, at the cost of fidelity.
On the other hand, the main effect of $\beta$ is registered on the size (right axis) that significantly grows with the number of input rules.
We remark that in this experiment we used $\alpha_q = 50$ that causes a smaller number of rules to be filtered out.
Therefore, the $\alpha$ ($\alpha_q$) parameter should be tuned not only according to the required fidelity and size, but also according to the number of input rules.
In the rest of this section we report results using all the local rules returned by \lore\ from $X_{le}$ if not differently specified.
\begin{table}[t!]
  \centering
  \footnotesize
  \setlength{\tabcolsep}{1mm}
  \begin{tabular}{cccccc|ccccccc}
    \toprule
     & \textbf{\textit{b}} &   \textbf{method} &  \textbf{fidelity} & \textbf{size} & \textbf{length} &  & \textbf{\textit{b}} &   \textbf{method} &  \textbf{fidelity} & \textbf{size} & \textbf{length} \\
    \midrule
    \parbox[t]{2mm}{\multirow{20}{*}{\rotatebox[origin=c]{90}{\adult, $\alpha = 10$}}}   & \multirow{6}{*}{DNN}    & \textsc{glx}  & .912              & \underline{5}   & $6.00 \pm 1.0$                &  \parbox[t]{2mm}{\multirow{20}{*}{\rotatebox[origin=c]{90}{\compas,  $\alpha = 5$}}}   & \multirow{6}{*}{DNN} & \textsc{glx}   &  .759              & \underline{3}   & \underline{$2.33 \pm 0.4$} \\
                                                                                                                  && \textsc{glx}*  & .880              & \underline{10}  & $6.30 \pm 2.5$                &                                                                                                             && \textsc{glx}*  &  .756              & \underline{6}   & $4.50 \pm 0.9$ \\
                                                                                                                  && \textsc{cpar}  & \underline{.929}  & 100             & $3.78 \pm 2.4$                &                                                                                                             && \textsc{cpar}  &  \underline{.821}  & 69              & $3.11 \pm 1.4$ \\
                                                                                                                  && \textsc{dt}    & .917              & 1068            & $7.22 \pm 1.9$                &                                                                                                             && \textsc{dt}    &  .789              & 1014            & $6.00 \pm 1.8$ \\
                                                                                                                  && \textsc{pdt}   & .908              & 28              & \underline{$2.71 \pm 0.8$}    &                                                                                                             && \textsc{pdt}   &  .780              & 30              & \underline{$2.33 \pm 0.6$} \\
                                                                                                                  && \textsc{uni}   & .880              & 6838            & $3.86 \pm 2.2$                &                                                                                                             && \textsc{uni}   &  .627              & 1515            & $4.65 \pm 1.7$ \\
    &&&&& & & &&&&& \\[-1ex]
                                                                                          & \multirow{6}{*}{RF}    &  \textsc{glx}  & .902              & \underline{10}  & $4.00 \pm 1.0$                &                                                                                     & \multirow{6}{*}{RF}    & \textsc{glx}  & .870                & \underline{6}   & $3.66 \pm 2.4$ \\
                                                                                                                  &&  \textsc{glx}* & .876              & \underline{10}  & $3.20 \pm 1.5$                &                                                                                                             && \textsc{glx}* & .862                & \underline{6}   & $4.33 \pm 1.1$ \\
                                                                                                                  &&  \textsc{cpar} & .944              & 107             & $2.57 \pm 1.3$                &                                                                                                             && \textsc{cpar} & \underline{.908}    & 53              & $2.86 \pm 1.6$ \\
                                                                                                                  &&  \textsc{dt}   & \underline{.959}  & 926             & $7.43 \pm 1.6$                &                                                                                                             && \textsc{dt}   & .906                & 452             & $4.91 \pm 1.4$ \\
                                                                                                                  &&  \textsc{pdt}  & .935              & 26              & \underline{$2.53 \pm 0.8$}    &                                                                                                             && \textsc{pdt}  & .886                & 30              & \underline{$2.60 \pm 0.7$} \\
                                                                                                                  &&  \textsc{uni}  & .876              & 838             & $5.00 \pm 1.4$                &                                                                                                             && \textsc{uni}  & .658                & 1515            & $3.52 \pm 1.2$ \\
    &&&&& & & &&&&& \\[-1ex]
                                                                                          & \multirow{6}{*}{SVM}   & \textsc{glx}   & .865              & \underline{10}  & $7.70 \pm 3.4$               &                                                                                      & \multirow{6}{*}{SVM}   & \textsc{glx}   & .860                & \underline{6}    & $4.33 \pm 1.3$ \\
                                                                                                                  && \textsc{glx}*  & .854              & \underline{10}  & $6.10 \pm 2.9$                &                                                                                                             && \textsc{glx}*  & .840               & \underline{6}   & $4.16 \pm 0.8$ \\
                                                                                                                  && \textsc{cpar}  & .848              & 95              & $4.77 \pm 3.0$                &                                                                                                             && \textsc{cpar}  & .858               & 70              & $3.04 \pm 1.3$ \\
                                                                                                                  && \textsc{dt}    & \underline{.875}  & 2956            & $7.57 \pm 1.6$                &                                                                                                             && \textsc{dt}    & \underline{.875}   & 824             & $4.58 \pm 1.0$ \\
                                                                                                                  && \textsc{pdt}   & .854              & 24              & \underline{$2.50 \pm 0.6$}    &                                                                                                             && \textsc{pdt}   & .850               & 30              & \underline{$2.53 \pm 0.7$} \\
                                                                                                                  && \textsc{uni}   & .854              & 6838            & $3.54 \pm 2.4$                &                                                                                                             && \textsc{uni}   & .696               & 1515            & $4.08 \pm 0.8$ \\
                              \midrule
    \parbox[t]{2mm}{\multirow{6}{*}{\rotatebox[origin=c]{90}{\diva, $\alpha = 25$}}}      & \multirow{6}{*}{RF}    & \textsc{glx}   & \underline{.854}  & \underline{26}   & $3.26 \pm 0.9$                &  \parbox[t]{2mm}{\multirow{6}{*}{\rotatebox[origin=c]{90}{\german,  $\alpha = 2$}}}& \multirow{6}{*}{RF}    & \textsc{glx}   & .786              & \underline{2}    & $3.00 \pm 1.0$ \\
                                                                                                                  && \textsc{glx}*  & .848              & \underline{26}   & $3.88 \pm 1.3$                &                                                                                                              && \textsc{glx}*  & .766              & \underline{2}    & $5.87 \pm 2.4$ \\
                                                                                                                  && \textsc{cpar}  & .850              & 221             & \underline{$2.03 \pm 1.0$}    &                                                                                                               && \textsc{cpar}  & .773               & 18               & \underline{$2.33 \pm 1.4$} \\
                                                                                                                  && \textsc{dt}    & .853              & 976             & $10.63 \pm 3.9$               &                                                                                                               && \textsc{dt}    & \underline{.830}   & 76               & $4.50 \pm 1.7$ \\
                                                                                                                  && \textsc{pdt}   & .836              & 28              & $3.21 \pm 0.9$                &                                                                                                               && \textsc{pdt}   & .796               & 28               & $2.78 \pm 0.8$ \\
                                                                                                                  && \textsc{uni}   & .794              & 2013            & $2.90 \pm 1.0$                &                                                                                                               && \textsc{uni}   & .766               & 210              & $5.62 \pm 2.4$ \\
    \bottomrule
  \end{tabular}
  \caption{Effectiveness of the global explainers in terms of fidelity, size and length on different datasets and black box classifiers $b$. \ethica\ and \ethica*\ (having not access to the data) are indicated with \textsc{glx} and \textsc{glx}*, respectively. For each dataset and black box the highest fidelity and lowest size and length are underlined. \ethica\ has a high fidelity comparable with the best performer, the lowest complexity in terms of size.}
  \label{tbl:ethica_vs_global_models}
\end{table}

\subsection{Effectiveness of the Global Explainers}
The final goal of \ethica\ is to provide a global explanation of a black box classifier.
In order to test the effectiveness in replicating the black box behavior we compare the fidelity and complexity of the explanation theories returned by \ethica\ and \ethica*\ with the rule sets returned by the interpretable global models (\textsc{dt}, \textsc{cpar}, \textsc{pdt}) trained on $X_{le}$ with the labels returned by the black box.
In addition, as a baseline local to global method we compare against an approach (\textsc{uni}) that simply performs the union of the local decision rule and adopts all of them as explanation theory.
For \ethica\ we select $\alpha$ to be lower than the smallest competitor, in this case \textsc{pdt}.
Table~\ref{tbl:ethica_vs_global_models} reports the results of this comparison.
\ethica\ and \ethica*\ are shortened as \textsc{glx} and \textsc{glx}* for readability purposes.
For each dataset and black box the highest fidelity and lowest size and length are underlined.

The results show the ability of \ethica\ to find explanation theories with a high fidelity and low complexity.
We observe that \ethica\ has a competing fidelity which is comparable to the one of the best global explainer (generally the \textsc{dt}) as it is always only less than a 0.1 lower.
The loss is more evident when explaining DNNs, while it is negligible for the other black box classifiers.
At the same time, \ethica\ or \ethica\* yield the lowest size (number of rules) of the global explanation resulting in a simple and compact but effective model.
It is important to notice that \ethica\ learns sets of rules one order of magnitude smaller than \cpar\ and one/two orders of magnitude smaller than the \textsc{dt}.
This suggests that accounting for complexity and fidelity at the same time in the merging process can yield good results in both metrics.
The \textsc{dt} has a comparable number of rules with \ethica\ and a fidelity within $8\%$ of the one of \ethica.
In terms of rule length, \ethica\ learns rules consistently shorter than the \textsc{dt}, but longer than \textsc{pdt} and \textsc{cpar} who are the best performers according to this evaluation metric. 
However, as previously stated, the sets of rules learned by \textsc{cpar} are one/two orders of magnitude larger than the size of the explanation theory returned by \ethica.
On the other hand, the \textsc{dt} has on average more than two times the number of rules of \ethica\ and in a real scenario this can cause confusion, especially if every feature model a complicated concept.

\ethica\ show remarkable higher performances than \textsc{uni}.
Also \ethica*, that does not have access to data, obtains similar fidelity to \ethica, with fidelities $3\%$ lower in the worst configuration.
This indicates that just aggregating all the local explanations together is not beneficial for obtaining an effective global explanation.
The local explanation rules must be carefully processed in order to remove useless and/or misleading local aspects that do not help in understanding the global reasons for the classification.
Finally, \ethica* shows a slightly lower performance than \ethica\, with fidelity scores lower than $2{-}3\%$ and non-monotonic pattern on length and size, which are shorter/longer \mbox{and smaller/larger across datasets and black boxes.}

We adopt the non-parametric Friedman test~\cite{richardson2010nonparametric} for comparing the average ranks of explanation methods over multiple datasets and black boxes with respect to the fidelity, size and length. 
The null hypothesis that all methods are equivalent is rejected for $\mathit{p{-}value} < 0.05$ for fidelity, $\mathit{p{-}value} < 0.0001$ for size, and $\mathit{p{-}value} < 0.0005$ for length.
\begin{table}[t!]
  \centering
  \footnotesize
  \begin{tabular}{cccccc}
    \toprule
    & \textbf{method} & \textbf{accuracy} & $\Delta_{acc}$ & \textbf{size} &  \textbf{length} \\ \midrule
    \parbox[t]{2mm}{\multirow{5}{*}{\rotatebox[origin=c]{90}{\adult}}}    & $b$ & $.801 \pm 0.056$  &&&\\ 
                                                                          & \textsc{cpar}      & $.816 \pm 0.148$   & $-0.091 \pm 0.042$      & $139.1 \pm 217.2$     & $7.526 \pm 1.464$ \\
                                                                          & \textsc{dt}        & $.945 \pm 0.062$   & $\;\;\;0.027 \pm 0.039$ & $1654.0 \pm 1134.5$   & $7.430 \pm 0.159$ \\
                                                                          & \textsc{glx}       & $.792 \pm 0.019$   & $0.101 \pm 0.013$       & $8.3 \pm 2.8$         & $5.533 \pm 1.601$ \\
                                                                          & \textsc{pdt}       & $.894 \pm 0.057$   & $-0.005 \pm 0.043$      & $16.6 \pm 14.4$       & $2.519 \pm 0.139$ \\
                                                                          &&&&&\\[-1ex]
    \parbox[t]{2mm}{\multirow{5}{*}{\rotatebox[origin=c]{90}{\compas}}}   & $b$ & $.673 \pm 0.033$ &&&\\
                                                                          & \textsc{cpar}      & $.790 \pm 0.108$   & $-0.072 \pm 0.377$      & $43.7 \pm 79.5$       & $4.948 \pm 1.306$ \\
                                                                          & \textsc{dt}        & $.593 \pm 0.514$   & $-0.264 \pm 0.366$      & $762.0 \pm 288.7$     & $4.988 \pm 0.225$ \\
                                                                          & \textsc{glx}       & $.726 \pm 0.002$   & $0.103 \pm 0.049$       & $5.00 \pm 1.73$       & $4.611 \pm 0.976$ \\
                                                                          & \textsc{pdt}       & $.868 \pm 0.025$   & $\;\;\;0.029 \pm 0.363$ & $20.0 \pm 17.3$       & $2.566 \pm 0.004$ \\
                                                                          &&&&&\\[-1ex]
    \parbox[t]{2mm}{\multirow{5}{*}{\rotatebox[origin=c]{90}{\diva}}}     & $b$ & $.908$ &&&\\
                                                                          & \textsc{cpar}      & $.850$             & $\;\;\;0.020$           & $221.0$               & $2.031$     \\ 
                                                                          & \textsc{dt}        & $.854$             & $-0.0170$               & $976.0$               & $10.680$    \\
                                                                          & \textsc{glx}       & $.824$             & $\;\;\;0.029$           & $26.0$                & $3.192$     \\ 
                                                                          & \textsc{pdt}       & $.836$             & $-0.021$                & $28.0$                & $3.214$     \\
                                                                          &&&&&\\[-1ex]
    \parbox[t]{2mm}{\multirow{5}{*}{\rotatebox[origin=c]{90}{\german}}}   & $b$ & $.700$&&&\\
                                                                          & \textsc{cpar}      & $.880$             & $-0.006$                & $5.2$                 & $2.103$ \\
                                                                          & \textsc{dt}        & $.915$             & $-0.000$                & $26.0$                & $2.789$ \\
                                                                          & \textsc{glx}       & $.726$             & $0.006$                 & $2.0$                 & $3.000$ \\
                                                                          & \textsc{pdt}       & $.898$             & $-0.000$                & $10.0$                & $1.892$ \\
    \bottomrule
  \end{tabular}
  \caption{Accuracy of the global explainers on $X_{\mathit{ts}}$ adopted as replacement of the black box models. For each dataset, the first line reports the average accuracy of the black box classifiers indicated with $b$. The average \textit{accuracy} (and standard deviation) among the various black box classifiers is reported for each global explainer. $\Delta_{acc}$ is calculated as the difference between the model fidelity and accuracy. The closer to $0$, the better it is. The average (and standard deviation) size and length complete the table.}
  \label{tbl:accuracy}
\end{table}

\subsection{Replacing the Black Box with a Global Explainer}
As final experiment, we test the performance of \ethica\ in terms of \textit{accuracy}, i.e., we consider predictions on the real dataset labels, rather than the ones predicted by the black box.
With this experiment, we aim to understand whether \ethica\ can be used to replace the black box classifier instead of being used only for understanding the classification reasons.
Table~\ref{tbl:accuracy} reports the average accuracy values across the various black box classifiers for \ethica\ and for the interpretable classifiers adopted as competitors on the held-out test set $X_{\mathit{ts}}$.
Table~\ref{tbl:accuracy} also reports the standard deviation of the accuracy and the \emph{accuracy delta} $\Delta_{acc} \in [0, 1]$ calculated as the difference between the model fidelity and accuracy.
An explainable model with minimum $\Delta_{acc}$ score ($\Delta_{acc} = 0$) is as accurate on the dataset labels as it is faithful to the black box labels. In other words, we can expect similar performances when the explainable model is deployed to predict the actual dataset labels.
As the fidelity-accuracy gap grows ($\Delta_{acc}$ approaching $1$), the explainable model is significantly more faithful to the black box and less accurate on the dataset, that is, it overfits the black box labels at the cost of the dataset labels. In other words, we should expect a worse performance when the explainable model is deployed the actual dataset labels.

The first line of each dataset reports the average accuracy for the black box classifiers $b$.

\begin{table}[t]
  \centering
  \begin{subtable}{0.45\linewidth}
    \footnotesize
    \centering
    \begin{tabular}{cc}
      \toprule
      & $\Delta_{acc}$ \\ \midrule
      \adult    & $0.128 \pm 0.008$ \\
      \compas   & $0.228 \pm 0.017$ \\
      \diva     & $0.026 \pm 0.023$ \\
      \german   & $0.009 \pm 0.007$ \\
      \bottomrule
    \end{tabular}
    \caption{Aggregation by dataset.}
  \end{subtable}%
  \begin{subtable}{0.45\linewidth}
    \footnotesize
    \centering
    \begin{tabular}{cc}
      \toprule
      & $\Delta_{acc}$ \\ \midrule
      \textsc{cpar}     & $-0.037 \pm 0.052$ \\
      \textsc{dt}       & $-0.063 \pm 0.135$ \\
      \textsc{glx}      & $0.073 \pm 0.015$ \\
      \textsc{pdt}      & $0.005 \pm 0.020$ \\
      \bottomrule
    \end{tabular}
    \caption{Aggregation by method.}
  \end{subtable}
\caption{Average $\Delta_{acc}$ on $X_{\mathit{ts}}$. The lower the value, the better are the performance of a global explainer as it indicates stability  between the capacity of mimicking the black box behavior and the ability of being adopted as a classifier.}
\label{tbl:accuracy_delta}
\end{table}

\ethica\ falls behind some of the competitors in terms of accuracy, with accuracy values lower up to $7\%$ and $15\%$ for \compas\ and \adult, respectively.
The $\Delta_{acc}$ is low for most models, with the \textsc{dt} showing a peculiar behavior.
Indeed, its $\Delta_{acc}$ is highly unstable, with values ranging from $-0.264$ to $0.027$.
\ethica, \textsc{cpar} and \textsc{pdt}, which all comprise of simpler models, show lower variance.
These results are strengthened by the numbers in Table~\ref{tbl:accuracy_delta} that report the average $\Delta_{acc}$ aggregated respectively per dataset and per global explanation model, i.e., a zoom out from the previous Table.
Datasets show a wildly different behavior, with $|\Delta_{acc}|$ as low as $0.006$ ($0.6\%$ absolute increase) and as high as $0.086$ ($9\%$ absolute increase).
Neither dataset size nor average fidelity or accuracy appear to correlate with these deltas.
Surprisingly, \textsc{pdt} shows a positive average $\Delta_{acc}$, indicating a better accuracy than fidelity.
While this confirms the above considerations on model simplicity, it should be noted that the \textsc{pdt} are actually showing better performance on a label distribution different than the one they were trained on.
Among the other models, \ethica\ shows the lowest $\Delta_{acc}$, $\approx \times 1.5$ times lower than \textsc{cpar} and $\approx 2.8\times$ lower than \textsc{dt}.
Therefore, since the explanation theory $E$ returned by \ethica\ guarantees not only high fidelity and small complexity, but also an high accuracy in classification, it can successfully be used for replacing the original black box classifier. 

\section{Conclusions}
\label{sec:conclusions}
In this paper we have proposed \ethica, a model agnostic \ltg\ explanation algorithm based on logic rules for AI systems using non interpretable machine learning models in the decision process.
Starting from local explanations, \ethica\ derives global explanations to describe the overall logic of a black box model.
The proposed method applies a hierarchical approach to derive a global explanation from the local ones.
\ethica\ tackles both explanation complexity and \textit{fidelity} in emulating the AI decision system behavior.
The results suggest that \ethica\ can be a valid Local to Global approach, as it tends to provide faithful and simple models.
\ethica\ outperforms the trivial union of rules and it is competitive with natively global explanators especially in terms of complexity.
Finally, experiments also highlight that the explanation theories of \ethica\ might be used directly as transparent predictors with performances similar to other global predictors.

The key advantage of \ethica\ lies in its flexibility: merging and regularizing explanations as they are generated allows for a plethora of extensions.
Among them we indicate direct human-guided regularization, with ad-hoc regularization penalizing reliance on some features rather than others, alignment to existing expert knowledge and balancing the fidelity-complexity equilibrium.
In this paper, we defined an explanation with logical rules.
A direct follow-up would be the extension to fuzzy and non-CNF rules, empowering reasoning with uncertainty.
Adaptation to non-logical domains such as sequences, text and images is a primary objective, either by mapping them to logical rules, or re-defining the merge and similarity function in different domains.
In addition, future investigations could be directed to the development of different merging functions and stopping criteria.
An obvious extension is to study how to consider non-logic explanations and the application to other families of black boxes.
Images and text may be good stride in this direction.
Finally, an interesting future research direction is to study how to provide more informative \textit{causal explanations}, able to capture the causal relationships among the (endogenous as well as exogenous) variables and the decision, based on data observed by appropriately querying the black box. 
  
\section*{Acknowledgment}
This work is partially supported by the European Community H2020 programme under the funding schemes: 
H2020-INFRAIA-2019-1: Research Infrastructure G.A. 871042 \emph{SoBigData++} (\href{http://www.sobigdata.eu}{\texttt{sobigdata.eu}}), 
G.A. 78835 \emph{Pro-Res} (\href{http://prores-project.eu/}{\texttt{prores.eu}}), 
G.A. 761758 \emph{Humane AI} (\href{https://www.humane-ai.eu/}{\texttt{humane-ai.eu}}), 
G.A. 825619 \emph{AI4EU} (\href{https://www.ai4eu.eu/}{\texttt{ai4eu.eu}}), and the ERC-2018-ADG G.A. 834756 ``XAI: Science and technology \mbox{for the eXplanation of AI decision making".}

\bibliographystyle{elsarticle-num}
\bibliography{biblio}

\begin{thebibliography}{10}
\expandafter\ifx\csname url\endcsname\relax
  \def\url#1{\texttt{#1}}\fi
\expandafter\ifx\csname urlprefix\endcsname\relax\def\urlprefix{URL }\fi
\expandafter\ifx\csname href\endcsname\relax
  \def\href#1#2{#2} \def\path#1{#1}\fi

\bibitem{pasquale2015black}
F.~Pasquale, The black box society: The secret algorithms that control money
  and information, Harvard University Press, 2015.

\bibitem{miller2019explanation}
T.~Miller, Explanation in artificial intelligence: Insights from the social
  sciences, Artificial Intelligence 267 (2019) 1--38.

\bibitem{freitas2014comprehensible}
A.~A. Freitas, Comprehensible classification models: a position paper, ACM
  SIGKDD explorations newsletter 15~(1) (2014) 1--10.

\bibitem{guidotti2018survey}
R.~Guidotti, A.~Monreale, S.~Ruggieri, F.~Turini, F.~Giannotti, D.~Pedreschi, A
  survey of methods for explaining black box models, ACM Computing Surveys
  (CSUR) 51~(5) (2018) 93.

\bibitem{propublica2013analysis}
L.~K. Jeff~Larson, Surya~Mattu, J.~Angwin, How we analyzed the compas
  recidivism algorithm (2013).

\bibitem{wachter2017right}
S.~Wachter, et~al., Why a right to explanation of automated decision-making
  does not exist in the general data protection regulation, Int. Data Privacy
  Law 7~(2) (2017) 76--99.

\bibitem{comande2017right}
G.~Malgieri, G.~Comand\'e, Why a right to legibility of automated
  decision-making exists in the {G}eneral {D}ata {P}rotection {R}egulation,
  Int. Data Privacy Law 7~(4) (2017) 243--265.

\bibitem{guidotti2018helping}
R.~Guidotti, et~al., Helping your docker images to spread based on explainable
  models, in: ECML-PKDD, Springer, 2018.

\bibitem{goebel2018explainable}
R.~Goebel, A.~Chander, K.~Holzinger, F.~Lecue, Z.~Akata, S.~Stumpf,
  P.~Kieseberg, A.~Holzinger, Explainable ai: the new 42?, in: International
  Cross-Domain Conference for Machine Learning and Knowledge Extraction,
  Springer, 2018, pp. 295--303.

\bibitem{adadi2018peeking}
A.~Adadi, M.~Berrada, Peeking inside the black-box: A survey on explainable
  artificial intelligence (xai), IEEE Access 6 (2018) 52138--52160.

\bibitem{ribeiro2016should}
M.~T. Ribeiro, S.~Singh, C.~Guestrin, Why should i trust you?: Explaining the
  predictions of any classifier, in: KDD, ACM, 2016, pp. 1135--1144.

\bibitem{ribeiro2018anchors}
M.~T. Ribeiro, S.~Singh, C.~Guestrin, Anchors: High-precision model-agnostic
  explanations, in: AAAI, 2018.

\bibitem{guidotti2018local}
R.~{Guidotti}, A.~{Monreale}, F.~{Giannotti}, D.~{Pedreschi}, S.~{Ruggieri},
  F.~{Turini}, Factual and counterfactual explanations for black box decision
  making, IEEE Intelligent Systems (2019) 1--1\href
  {http://dx.doi.org/10.1109/MIS.2019.2957223}
  {\path{doi:10.1109/MIS.2019.2957223}}.

\bibitem{panigutti2019marlena}
C.~Panigutti, et~al., Explaining multi-label black box classifiers for health
  applications, in: W3PHIAI, Springer, 2019.

\bibitem{craven1996extracting}
M.~Craven, J.~W. Shavlik, Extracting tree-structured representations of trained
  networks, in: NIPS, 1996, pp. 24--30.

\bibitem{deng2014interpreting}
H.~Deng, Interpreting tree ensembles with intrees, arXiv preprint
  arXiv:1408.5456.

\bibitem{lundberg2017unified}
S.~M. Lundberg, S.-I. Lee, A unified approach to interpreting model
  predictions, in: Advances in neural information processing systems, 2017, pp.
  4765--4774.

\bibitem{yin2003cpar}
X.~Yin, J.~Han, Cpar: Classification based on predictive association rules, in:
  Proceedings of the 2003 SIAM International Conference on Data Mining, SIAM,
  2003, pp. 331--335.

\bibitem{lakkaraju2016interpretable}
H.~Lakkaraju, et~al., Interpretable decision sets: A joint framework for
  description and prediction, in: KDD, ACM, 2016, pp. 1675--1684.

\bibitem{ross2018learning}
A.~S. Ross, W.~Pan, F.~Doshi-Velez, Learning qualitatively diverse and
  interpretable rules for classification, arXiv preprint arXiv:1806.08716.

\bibitem{ross2019ensembles}
A.~S. Ross, W.~Pan, L.~A. Celi, F.~Doshi-Velez, Ensembles of locally
  independent prediction models, arXiv preprint arXiv:1911.01291.

\bibitem{pedreschi2019meaningful}
D.~Pedreschi, F.~Giannotti, R.~Guidotti, A.~Monreale, S.~Ruggieri, F.~Turini,
  Meaningful explanations of black box ai decision systems, in: Proceedings of
  the AAAI Conference on Artificial Intelligence, Vol.~33, 2019, pp.
  9780--9784.

\bibitem{guidotti2019factual}
R.~Guidotti, A.~Monreale, F.~Giannotti, D.~Pedreschi, S.~Ruggieri, F.~Turini,
  Factual and counterfactual explanations for black box decision making, IEEE
  Intelligent Systems.

\bibitem{alvarez2018robustness}
D.~Alvarez-Melis, T.~S. Jaakkola, On the robustness of interpretability
  methods, arXiv preprint arXiv:1806.08049.

\bibitem{guidotti2019stability}
R.~Guidotti, S.~Ruggieri, On the stability of interpretable models, in: 2019
  International Joint Conference on Neural Networks (IJCNN), IEEE, 2019, pp.
  1--8.

\bibitem{quinlan1993c4}
J.~R. Quinlan, C4. 5: Programs for Machine Learning, Elsevier, 1993.

\bibitem{yoon2012classification}
J.~Yoon, D.-W. Kim, Classification based on predictive association rules of
  incomplete data, IEICE TRANSACTIONS on Information and Systems 95~(5) (2012)
  1531--1535.

\bibitem{schmidhuber2015deep}
J.~Schmidhuber, Deep learning in neural networks: An overview, Neural networks
  61 (2015) 85--117.

\bibitem{lakkaraju2017interpretable}
H.~Lakkaraju, E.~Kamar, R.~Caruana, J.~Leskovec, Interpretable \& explorable
  approximations of black box models, arXiv preprint arXiv:1707.01154.

\bibitem{angelino2017learning}
E.~Angelino, N.~Larus-Stone, D.~Alabi, M.~Seltzer, C.~Rudin, Learning
  certifiably optimal rule lists, in: KDD, ACM, 2017, pp. 35--44.

\bibitem{ruggieri2004yadt}
S.~Ruggieri, Yadt: Yet another decision tree builder, in: Tools with Artificial
  Intelligence, ICTAI., IEEE, 2004, pp. 260--265.

\bibitem{craven1994using}
M.~W. Craven, J.~W. Shavlik, Using sampling and queries to extract rules from
  trained neural networks, in: JMLR, Elsevier, 1994, pp. 37--45.

\bibitem{tan2007introduction}
P.-N. Tan, et~al., Introduction to data mining, Pearson Education India, 2007.

\bibitem{DBLP:conf/nips/LundbergL17}
S.~M. Lundberg, S.~Lee, A unified approach to interpreting model predictions,
  in: I.~Guyon, U.~von Luxburg, S.~Bengio, H.~M. Wallach, R.~Fergus, S.~V.~N.
  Vishwanathan, R.~Garnett (Eds.), Advances in Neural Information Processing
  Systems 30: Annual Conference on Neural Information Processing Systems 2017,
  4-9 December 2017, Long Beach, CA, {USA}, 2017, pp. 4765--4774.

\bibitem{shapley1953value}
L.~S. Shapley, A value for n-person games, Contributions to the Theory of Games
  2~(28) (1953) 307--317.

\bibitem{rokach2005clustering}
L.~Rokach, O.~Maimon, Clustering methods, in: Data mining and knowledge
  discovery handbook, Springer, 2005, pp. 321--352.

\bibitem{wit2012all}
E.~Wit, E.~v.~d. Heuvel, J.-W. Romeijn, ‘all models are wrong...’: an
  introduction to model uncertainty, Statistica Neerlandica 66~(3) (2012)
  217--236.

\bibitem{pelleg2000x}
D.~Pelleg, A.~W. Moore, et~al., X-means: Extending k-means with efficient
  estimation of the number of clusters., in: Icml, Vol.~1, 2000, pp. 727--734.

\bibitem{guidotti2015tosca}
R.~Guidotti, R.~Trasarti, M.~Nanni, Tosca: two-steps clustering algorithm for
  personal locations detection, in: Proceedings of the 23rd SIGSPATIAL
  International Conference on Advances in Geographic Information Systems, 2015,
  pp. 1--10.

\bibitem{DBLP:journals/ml/Furnkranz97}
J.~F{\"{u}}rnkranz, \href{https://doi.org/10.1023/A:1007329424533}{Pruning
  algorithms for rule learning}, Machine Learning 27~(2) (1997) 139--172.
\newblock \href {http://dx.doi.org/10.1023/A:1007329424533}
  {\path{doi:10.1023/A:1007329424533}}.
\newline\urlprefix\url{https://doi.org/10.1023/A:1007329424533}

\bibitem{ruggieri2013learning}
S.~Ruggieri, Learning from polyhedral sets (2013).

\bibitem{bertsimas2017optimal}
D.~Bertsimas, J.~Dunn, Optimal classification trees, Machine Learning 106~(7)
  (2017) 1039--1082.

\bibitem{richardson2010nonparametric}
A.~Richardson, Nonparametric statistics for non-statisticians: a step-by-step
  approach by gregory w. corder, dale i. foreman, International Statistical
  Review 78~(3) (2010) 451--452.

\end{thebibliography}

\end{document}